\documentclass{article}

 \usepackage[preprint]{neurips_2026}

\usepackage[utf8]{inputenc} 
\usepackage[T1]{fontenc}    
\usepackage{url}            
\usepackage{booktabs}       
\usepackage{amsfonts}       
\usepackage{nicefrac}       
\usepackage{microtype}      

\usepackage{amsmath}
\usepackage{amsthm}
\newtheoremstyle{italichead}
  {}{}                  
  {\normalfont}         
  {}                    
  {\itshape}            
  {.}                   
  { }                   
  {\thmname{#1}\thmnumber{ #2}\thmnote{ (#3)}}
\theoremstyle{italichead}
\newtheorem{proposition}{Proposition}
\newtheorem{corollary}{Corollary}

\usepackage[table]{xcolor}
\usepackage{graphicx}
\usepackage{enumitem}
\usepackage{graphicx}
\usepackage{subcaption}
\usepackage{fontawesome5}
\usepackage{setspace}

\usepackage[colorlinks=true,citecolor=blue,linkcolor=blue,urlcolor=blue]{hyperref}
\usepackage{bm}
\usepackage{multirow}
\usepackage{tabularx}
\usepackage{array}
\definecolor{citeblue}{RGB}{90,130,180}
\definecolor{lightblue}{RGB}{232,241,252}
\definecolor{lightgreen}{RGB}{235,247,235}
\definecolor{termorange}{RGB}{210,105,30}
\definecolor{termblue}{RGB}{30,100,180}
\hypersetup{
  colorlinks=true,
  citecolor=citeblue,
  linkcolor=red,
  urlcolor=purple
}

\newcommand{\Ours}{\textsc{Target-SFT}}
\newcommand{\ours}{\textsc{Target}}
\newcommand{\projecturl}{https://txie1.github.io/Target-SFT/}

\newcolumntype{Y}{>{\raggedright\arraybackslash}X}
\newcolumntype{M}{>{\centering\arraybackslash}m{0.38\textwidth}}
\newcommand{\lightrule}{\arrayrulecolor{black!20}\midrule\arrayrulecolor{black}}

\title{A Unifying Lens on Supervised Fine-Tuning Through Target Distribution Design
}

\author{%
  Tong Xie$^{1}$ \hspace{0.2em}
  Yuanhao Ban$^{1,2}$ \hspace{0.2em}
  Yunqi Hong$^{1}$ \hspace{0.2em}
  Sohyun An$^{1}$ \hspace{0.2em}
  Yihang Chen$^{1}$ \hspace{0.2em}
  Cho-Jui Hsieh$^{1,2}$
  \\[0.4em]
  $^{1}$University of California, Los Angeles (UCLA), 
  \hspace{0.5em}
  $^{2}$Arena
  \\[0.2em]
  \texttt{\{tongxie,chohsieh\}@cs.ucla.edu}
  \\[0.6em]
  \textbf{Project Page: \href{\projecturl}{\faIcon{globe}~Target-SFT}}
}

\begin{document}

\maketitle


\begin{abstract}

Supervised fine-tuning (SFT) typically maximizes the likelihood of every token in a demonstrated trajectory. However, an observed token can be non-unique, noisy, or misaligned with the model prior. Strictly fitting toward this one-hot target may be suboptimal, especially when the pretrained model encodes a rich knowledge prior. In this work, we reinterpret SFT as target distribution design: instead of studying only the loss objective, we analyze the token-level target that the loss drives the model to match. We introduce the $Q$-target framework, which decomposes SFT supervision into two explicit choices: (1) \textbf{how strongly to rely on the observed token}, and (2) \textbf{how to allocate the remaining probability mass} over alternatives. This perspective unifies many existing SFT variants as implicit choices of the target distribution $Q$. Building on this view, we propose \Ours{} which constructs the training objective directly from the desired target distribution. This method consistently outperforms across the ten reasoning dataset-model settings evaluated, showing the effectiveness of this target-based approach. Overall, our formulation reveals a more fundamental design principle for SFT training and opens a broader search space for SFT objectives.
\end{abstract}
\section{Introduction}

Supervised fine-tuning (SFT) is a central stage in the post-training of large language models (LLMs)~\cite{zhang2025, chung2022, ouyang2022}. By imitating expert behaviors, SFT enables the pretrained model to acquire new knowledge and adapt to tasks efficiently. Despite its popularity, standard SFT relies on a particularly rigid form of supervision, by training the model toward a one-hot target distribution: at every token position, SFT maximizes the probability of the demonstrated token $y_t$, while all other tokens are assigned zero probability. This formulation implicitly assumes that every observed token in the dataset is an ideal and uniquely correct target.

This one-hot view reveals a limitation of standard SFT, especially in post-training settings~\cite{falsepromise, gem, beyondlog}. In realistic SFT data, an observed token is rarely the only valid continuation. The same prompt may admit multiple correct reasoning paths, phrasings, intermediate steps, or stylistic choices~\cite{surveydataselection, surveydataforft, rft, lima, profit, 8020}. At the same time, the model already encodes a rich prior from pretraining~\cite{beyondlog, dft, profit}. In such cases, forcing the model to strictly imitate every token can amplify noise, induce overconfidence, interfere with the pretrained model prior, and degrade generalization~\cite{sftmemorize, rlsrazor, retainbydoing, sftlabelsmooth, grape}. A growing line of work relaxes the SFT supervision by modifying the objective, for example, through token-level importance reweighting~\cite{dft, profit, cft, eaft} or regularization~\cite{gem, sftlabelsmooth, asft, proximalsft}. While these approaches are effective, they are often presented as separate algorithmic choices. It remains unclear the connection between variants and how to construct better SFT objectives.

\begin{figure}[t]
  \centering
  \includegraphics[width=\linewidth]{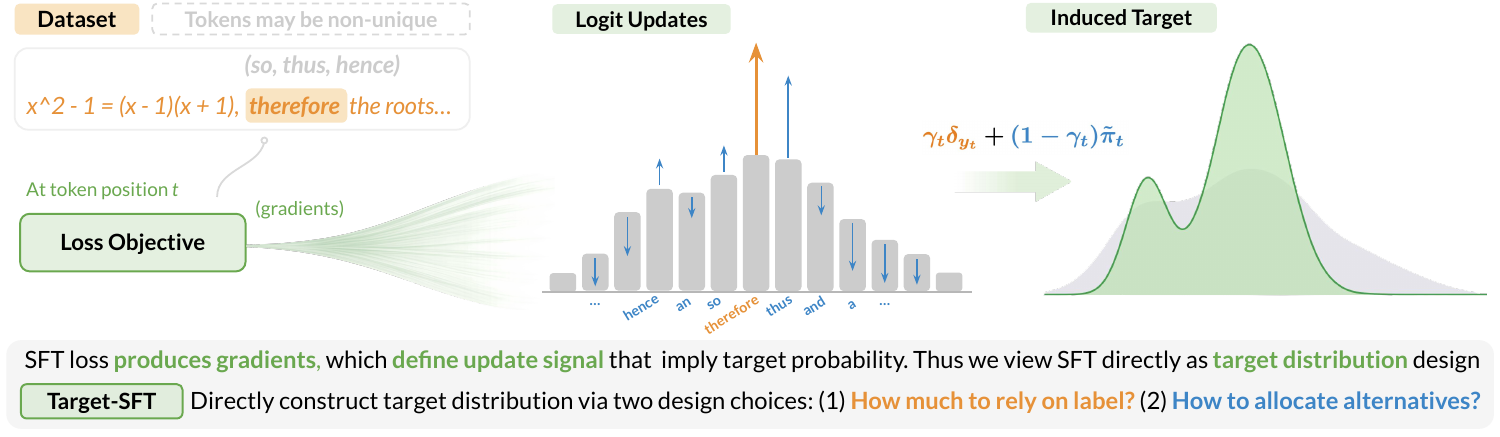}
  \caption{\textbf{Overview.} An SFT loss drives the model to match an implicitly defined target distribution. This view motivates \Ours{} that designs the SFT target directly. It also offers a unifying lens, where many SFT variants can be viewed as different target designs through the choices of $\gamma_t$ and $\tilde{\pi}_t$.}
  \label{fig:overview}
\end{figure}

In this work, we propose a different perspective: rather than the loss, we ask what \textbf{target distribution} should SFT drive the model to learn. This is more fundamental than the choice of loss alone, because loss is merely an optimization surrogate, while the target distribution directly specifies the desired allocation of probability mass (Figure~\ref{fig:overview}). By viewing SFT as target distribution design, we can control the supervision signal when the observed label $y_t$ is suboptimal: If $y_t$ is ideal and unique, the target should be close to the one-hot distribution $\delta_{y_t}$, maximizing its probability. If it is noisy or misaligned with the model prior, then the target should soften supervision and allocate probability to alternatives. Building on this intuition, we introduce a $Q$-target distribution framework for SFT:
\begin{align*}
    Q_t=\gamma_t\delta_{y_t}+(1-\gamma_t)\tilde{\pi}_t,
\end{align*}
where $\gamma_t\in[0,1]$ controls the target probability assigned to the observed token $y_t$ based on uncertainty, and $\tilde{\pi}$ specifies the plausible alternatives for the $1-\gamma_t$ residual probability mass. This perspective thus reveals two key questions through the choices of $(\gamma_t, \tilde{\pi}_t)$: (1) \textbf{how much to rely on the observed token $y_t$}, and (2) \textbf{where to allocate the remaining probability mass when $y_t$ is uncertain}?

In particular, we show that many existing SFT variants can be understood as varying ways of answering these two questions, and seemingly different losses correspond to implicit choices of target distribution $Q$. Based on this insight, we propose \Ours{} by explicitly leveraging the structure revealed by the $Q$-framework, which previous methods have largely left implicit.
In general, we argue that the fundamental object in SFT is not the loss function itself, but the target distribution induced by the loss. This $Q$-target perspective provides a unifying lens in SFT objective design, and exposes a general design space for balancing dataset imitation, prior preservation, and alternative supervision. Our contributions are as follows:
\begin{enumerate}[leftmargin=2.5em, itemsep=1pt]
    \item We introduce a target-distribution perspective on SFT, showing that arbitrary token-level SFT losses can be understood through the induced target distributions they drive the model to learn.
    \item We propose the $Q$-target framework, which unifies existing SFT variants by decomposing objective design into two explicit choices: how much to rely on the observed token and how to allocate the remaining probability mass.
    \item We propose \Ours{} as a concrete instantiation motivated by the $Q$-target view, and empirically validate its performance across 10 dataset-model settings.
\end{enumerate}

\section{Related Work}
\label{sec:related}

Existing works improve SFT along three main directions. We organize them under the $Q$-target lens, which characterizes SFT variants by how they specify the effective token-level target distribution. We include the concrete connection for representative methods in Appendix~\ref{app:unify}.

\paragraph{Token-level Reweighting.}
Standard SFT applies uniform cross-entropy updates to all tokens, treating every token as equally reliable. Token-reweighting methods challenge this assumption by changing how strongly each token contributes to training. DFT~\cite{dft}, beyond-log~\cite{beyondlog}, and ProFit~\cite{profit} use the model probability on the observed token to rescale or filter updates, focusing on tokens that are compatible with the model prior. EAFT~\cite{eaft} uses entropy-based uncertainty to reduce potentially destructive updates; iw-SFT~\cite{iwsft} and CFT~\cite{cft} assign weights based on trajectory- or token-level quality. These methods primarily address the \textit{choice of $\gamma_t$ in SFT target construction}, controlling how much target mass should be assigned to the observed label. However, reweighting the one-hot loss only weakens or strengthens imitation, but leaves the remaining probability mass underspecified. Our framework provides a complete view by making the effective training target explicit. 

\paragraph{Distribution-level Prior.}
Another line of work introduces soft distributional signals beyond the one-hot label. Reference-constrained methods such as ASFT~\cite{asft}, RL’s Razor~\cite{rlsrazor}, and Proximal SFT~\cite{proximalsft} regularize updates to prevent large drift from a reference model. Huang et al.~\cite{sftlabelsmooth} uses label-smoothing to address overconfidence. GEM~\cite{gem} uses reverse KL and entropy regularization to preserve output diversity and reduce forgetting. These methods address the rigidity in strict one-hot imitation, proposing alternative sources for probability allocation. They mainly specify the \textit{choice of $\tilde{\pi}$, allocating probability mass to non-observed alternatives}: KL-constraints allocate toward the reference model, label smoothing allocates toward a uniform prior, and diversity-preserving methods discourage collapse onto a narrow set of tokens. 

\paragraph{Dataset-level Curations.}
A complementary direction improves SFT by changing the training trajectories. Prior work has proposed augmenting demonstrations with multiple valid trajectories~\cite{rft, metamath}, filtering examples by quality~\cite{alpagasus, rest}, and using model-generated or model-aligned responses~\cite{retainbydoing}. GRAPE~\cite{grape} selects trajectories with high likelihood under the model, while rejection-sampling fine-tuning trains on correct model-generated responses~\cite{rft, selfplay, star, raft}. Self-distillation fine-tuning~\cite{selfdistillft} projects expert demonstrations into the model’s distributional style, reducing data-model mismatch. These approaches address the same underlying issue as our work: the demonstrated trajectory or token may not be uniquely ideal for the model to imitate. By improving the dataset, they indirectly change the effective target distribution seen during SFT. In contrast, our work remains on the objective level and directly designs the effective target distribution for a fixed dataset.
\section{Preliminary}

\paragraph{Supervised Fine-Tuning.}

Let $\mathcal{D}$ be a supervised dataset of pairs $(x,y)\sim\mathcal{D}$, where $x$ is the input prompt and $y=(y_1,\dots,y_T)$ is the demonstrated response sequence. Given the prefix $x_t=(x,y_{<t})$, a language model defines the next-token distribution $\pi_\theta(\cdot \mid x_t)\in \Delta^{|\mathcal{V}|}$ over the vocabulary $\mathcal{V}$.

Standard SFT minimizes the token-level negative log-likelihood:
\begin{align*}
\mathcal{L}_{\mathrm{SFT}}(\theta)
=
\mathbb{E}_{(x,y)\sim\mathcal{D}}
\left[
-\sum_{t=1}^T \log \pi_\theta(y_t\mid x_t)
\right].
\end{align*}

\paragraph{Target Distribution.}
Equivalently, let $\delta_{y_t}$ denote the one-hot vector that assigns probability $1$ to the observed token $y_t$ and $0$ to all other tokens, $\delta_{y_t}(v)=\mathbf{1}\{v=y_t\}$. Then the objective can be written as the cross-entropy with target $\delta_{y_t}$ as $\mathcal{L}_{\mathrm{SFT}}(\theta)
=
\mathbb{E}_{(x,y)\sim\mathcal{D}}
\sum_{t=1}^{T}
\mathrm{CE}\bigl(\delta_{y_t},\pi_\theta(\cdot\mid x_t)\bigr).$

\section{Q-Target Framework for SFT}

The one-hot target $\delta_y$ in SFT implicitly assumes that $y_t$ is the single optimal continuation for the prefix $x_t$. However, an observed token can be non-unique, noisy, or distributionally mismatched with the model prior. To capture this, we relax the assumption and account for the uncertainty in $y_t$, constructing a new target distribution $Q_t$ in place of $\delta_y$.

\subsection{Modeling Latent Trust}

We introduce a latent binary variable to represent whether the observed token should be strictly imitated. Let $Z_t \in \{0,1\}$ be where $Z_t=1$ indicates $y_t$ is strictly trusted as the target, and $Z_t=0$ indicates that supervision should relax to a broader distribution over plausible alternatives. Under this view, the ideal target distribution can be written as
\begin{align}
\label{eq:Z}
    P(\cdot \mid x_t) = P(Z_t=1 \mid x_t)\delta_{y_t} + P(Z_t=0 \mid x_t)\tilde{\pi}_t(\cdot \mid x_t),
\end{align}
where $\tilde{\pi}_t \in \Delta^{|\mathcal{V}|}$ denotes an alternative distribution over plausible next tokens.

Since $Z_t$ is unobserved, the trust probability $r_t = P(Z_t=1 \mid x_t)$ is unknown. We model this uncertainty using a Beta distribution:
\begin{align*}
    r_t \sim \mathrm{Beta}(\alpha_t, \beta_t),
\end{align*}
where $\alpha_t$ is evidence supporting $y_t$, and $\beta_t$ is evidence that $y_t$ may be non-unique or should relax toward alternatives. The posterior mean $\gamma_t = \mathbb{E}[r_t] = \frac{\alpha_t}{\alpha_t + \beta_t} \in [0,1]$ then gives the expected trust for the observed token $y_t$. 

Taking expectation over Eq.~\eqref{eq:Z} leads to the ideal target distribution
\begin{align}
\label{eq:Q_as_expectation}
    Q_t = \mathbb{E}_{r_t}\!\left[P(\cdot\mid x_t)\right]
    = \gamma_t\delta_{y_t} + (1-\gamma_t)\tilde{\pi}_t.
\end{align}
Intuitively, the target probability for $y_t$ is scaled by the expected trust $\gamma_t$ in the token, and the residual probability mass is reallocated to plausible alternatives over $\tilde{\pi}_t$.

\subsection{Final Target \& Objective}

We replace the SFT one-hot target with $Q_t$, and train with the cross-entropy loss:
\begin{align*}
\mathcal{L}_{Q}(\theta)
=
\mathbb{E}_{(x,y)\sim\mathcal{D}}
\sum_{t=1}^{T}
\mathrm{CE}\!\left(Q_t,\pi_\theta(\cdot\mid x_t)\right). 
\end{align*}
\begin{proposition}[Token-level decomposition of $Q$-target training]
\label{prop:q_decomp}

Given the target distribution $Q_t$ defined in Eq.~\eqref{eq:Q_as_expectation}, the token-level cross-entropy loss at position $t$ decomposes as
\begin{align}
    \mathrm{CE}\!\left(Q_t,\pi_\theta(\cdot\mid x_t)\right)
=
\gamma_t\,\textcolor{termorange}{\mathrm{CE}\!\left(\delta_{y_t},\pi_\theta(\cdot\mid x_t)\right)}
+
(1-\gamma_t)\,
\textcolor{termblue}{\mathrm{CE}\!\left(\tilde{\pi}_t,\pi_\theta(\cdot\mid x_t)\right)}.
\label{eq:q_decomposition}
\end{align}
\end{proposition}

This shows that training to match the $Q$-target involves two forms of supervision: (1) \textcolor{termorange}{label imitation}, which pushes the model toward the observed token $y_t$, with strength controlled by the expected trust $\gamma_t$, and (2)
\textcolor{termblue}{residual distribution matching}, which assigns the remaining supervision mass to alternatives through $\tilde{\pi}_t$. See Appendix~\ref{app:proof_loss_decomp} for proof.

\section{Unifying Perspective}

\subsection{Existing Variants}
\label{sec:unifying}

The $Q$-target formulation separates token-level supervision into two design choices: (1) $\gamma_t \in [0,1]$ controls the target probability mass on $y_t$, while (2) $\tilde{\pi}_t\in\Delta^{|\mathcal V|}$ specifies how the residual probability is allocated. We show that this view unifies many existing SFT variants. Table~\ref{tab:sft_variants} provides details of each variant discussed, and Table~\ref{tab:unify} summarizes their interpretation under $Q$-target framework.

\paragraph{Standard SFT.}
Consider the degenerate choice $\gamma_t=1$, then the $Q$-objective in Eq.~\eqref{eq:q_decomposition} reduces to the negative log-likelihood in SFT, corresponding to setting $Q_{t,k}=\delta_{y_t}$:
\begin{align*}
\mathrm{CE}(Q_t,\pi_\theta)
&=\mathrm{CE}(\delta_{y_t},\pi_\theta)
=-\log \pi_\theta(y_t\mid x_t),
&
Q_t(k)
&=\delta_{y_t}(k)
=\begin{cases}
1, & k = y_t,\\
0, & k \neq y_t.
\end{cases}
\end{align*}
Hence, standard SFT is the special case that places full probability on every observed token $y_t$ and assigns no residual mass to alternatives.

\paragraph{Token-Weighted Variants.} 
A class of SFT variants modifies the objective by scaling the negative log-likelihood with a detached, per-token importance weight $w_t$~\cite{beyondlog, profit, dft, cft, eaft, iwsft}:
\begin{align*}
    \mathcal{L}_t = -w_t\log \pi_\theta(y_t\mid x_t),
\end{align*}
where $w_t$ may depend on model confidence, entropy, sample quality, or other token-level statistics.

\begin{corollary}[Token weighting as self-residual $Q$-target.] 
\label{cor:token_variant}
Assume $w_t\in[0,1]$ is detached from the current update. The token-weighted loss above corresponds to the choice
\begin{align*}
\left(
\gamma_t=w_t,\,
\tilde{\pi}_t=\mathrm{sg}\!\left[\pi_\theta(\cdot\mid x_t)\right]
\right)
\quad\Longrightarrow\quad
Q_t = w_t\delta_{y_t} + (1-w_t)\mathrm{sg}\!\left[\pi_\theta(\cdot\mid x_t)\right].
\end{align*}
where $\mathrm{sg}[\cdot]$ denotes the stop-gradient operator. In particular, the residual branch $\tilde{\pi}_t$ is a \textit{self-matching} term that contributes no gradient. See Appendix~\ref{app:proof_token_variant} for proof.
\end{corollary}
This shows that token-weighted variants primarily specify $\gamma_t$, proposing statistics to determine how strongly to imitate the observed token. And the residual mass $1-\gamma_t$ is allocated to the current model prior $\mathrm{sg}\!\left[\pi_\theta(\cdot\mid x_t)\right]$, providing no corrective supervision toward potential alternatives.


\paragraph{Distributional Variants.}
Another class of SFT variants incorporates distributional signals beyond the observed token~\cite{gem, sftlabelsmooth, asft, proximalsft, minillm}. With various intended goals (e.g., to regularize model drift, calibrate confidence, preserve output diversity, etc), these methods enrich hard-label imitation using another distribution target $q_t$. The objective is of the form
\begin{align}
    \mathcal{L}_t
    =
    -a_t \log \pi_\theta(y_t\mid x_t)
    +
    b_t\,\mathrm{CE}(q_t,\pi_\theta),
    \qquad a_t,b_t\ge 0.
    \label{eq:distributional_objective}
\end{align}

\begin{corollary}[Distributional variants as residual Q-targets]
\label{cor:dist_variant}
Given a detached, auxiliary or reference distribution $q_t\in\Delta^{|\mathcal V|-1}$, distributional variants correspond to $Q$-target training with
\begin{align}
    \gamma_t
    =
    \frac{a_t}{a_t+b_t},
    \qquad
    \tilde{\pi}_t=q_t,
    \quad\Longrightarrow\quad
    Q_t
    =
    \gamma_t\delta_{y_t}
    +
    (1-\gamma_t)\tilde{\pi}_t,
    \label{eq:distributional_q_target}
\end{align}
up to a global constant $a_t+b_t$. Therefore, these methods primarily specify the residual branch $\tilde{\pi}_t$, deciding where non-label probability should be allocated. The relative strength $\gamma_t$ between label imitation and residual matching is determined by fixed hyperparameter. See Appendix~\ref{app:proof_dist_variant} for proof.
\end{corollary}

In summary, token-weighted variants mainly design the label-trust coefficient $\gamma_t$, while distributional variants design the residual distribution $\tilde{\pi}_t$. 
Together, they present the two axes in our framework: how strongly to imitate the observed token, and how to allocate the remaining probability mass.

\paragraph{Remark.}
This is a natural view because an SFT loss is a training surrogate. Although variant losses may take different algebraic forms, their effect on the model is mediated through the probability update over the vocabulary. The loss expression is therefore only a way to generate gradients; the induced target distribution reveals what those gradients effectively drive the model to match. In this sense, the $Q$-target formulation is a more fundamental perspective beyond the loss forms. This view not only unifies variants but provides a direct lens into training signals. We now formalize this idea, where we derive the induced target $Q_t$ for any arbitrary differentiable token-level loss.

\subsection{From Any Loss to $Q$}
\label{sec:any_loss_to_Q}

An SFT loss defines a surrogate for shaping the model's next-token distribution. At each prefix $x_t$, the model outputs a distribution $p_t=\pi_\theta(\cdot\mid x_t)$, and the loss produces gradients that determine how this probability changes across tokens. Therefore, for any differentiable token-level loss, we can ask \textit{what target probability distribution $Q_t$ does the loss drive the model to match through its gradients}?

Given token position $t$ (notation omitted for clarity), consider the cross-entropy toward target $Q_t$
\begin{align*}
    \mathcal{L}_{\mathrm{CE}}(Q,p)
=
-\sum_{k\in\mathcal{V}} Q_k \log p_k.
\end{align*}
Let $z$ denote the logits. The gradient with respect to the $k$-th logit is simply the prediction difference
\begin{align*}
    \frac{\partial \mathcal{L}_{\mathrm{CE}}}{\partial z_k}
=
p_k-Q_k.
\end{align*}
This relationship can be inverted. Given any differentiable token-level loss $\mathcal{L}(z,x)$ with logit gradient $g_k = \frac{\partial \mathcal{L}}{\partial z_k}$, we can derive its induced target as
\begin{align}
\label{eq:induced_Q}
    Q_k
:=
p_k-g_k.
\end{align}
This $Q$ is the target whose cross-entropy gradient $\nabla_z \mathrm{CE}(Q,p)=\nabla_z \mathcal{L}$, aligning exactly with the logit updates produced by $\mathcal{L}$. It therefore explicitly reveals the training signal encoded by the loss. Appendix~\ref{app:loss-to-q} shows example derivations, and visualizes loss through its gradients and target $Q$.

\section{\Ours}

The $Q$-formulation turns SFT supervision from a static log-likelihood objective into a problem of target distribution design: (1) how much to rely on the observed token $y_t$, and (2) how to allocate the remaining probability mass.  We now introduce \Ours{}, which leverages both branches of this construction. We first use a model-based proxy to estimate label uncertainty, and motivate for an external teacher distribution to enrich supervision signals through the residual branch. 

\paragraph{Probability-Proxy for $\bm{\gamma_t}$.}
The ideal target in Eq~\eqref{eq:Q_as_expectation} involves an expected trust $\gamma_t = \mathbb{E}[r_t] = \frac{\alpha_t}{\alpha_t + \beta_t}$, where $\alpha_t, \beta_t$ represent evidence (such as an empirical count) for the binary event $y_t$ being selected given prefix $x_t$. However, such a count is intractable in SFT.

Instead, the model probability $p_y=\pi_\theta(y_t\mid x_t)$ arises as a natural proxy for $\alpha_t$, which encapsulates statistical evidence accumulated during pretraining. Here $p_y$ represents the fraction of the model's belief assigned to $y_t$ among all possible continuations. By defining the evidence as $\alpha_t=p_y$ and $\beta_t=1-p_y$, the posterior mean resolves to
\begin{align*}
\gamma_t &= \frac{p_y}{p_y + (1 - p_y)} = p_y.
\end{align*}
This derivation motivates probability-weighted SFT variants (such as $p$-loss)~\cite{dft, beyondlog}, as scaling the SFT target by $p_y$ equates to using predictive probability as proxy measure of uncertainty in the label.

\paragraph{Teacher-Guided Reward Shaping $\bm{\tilde{\pi}}$.}

Under the $Q$-target view, such objectives implicitly use a \textit{self-matching} residual, $\tilde{\pi}_t=\mathrm{sg}\!\left[\pi_\theta(\cdot\mid x_t)\right],$ as derived in Section~\ref{sec:unifying}. This shows that as $p_y \rightarrow 0$, the supervision weakens and provides no further corrective gradient. This is limiting because a small $p_y$ may arise from uncertainty in label, or indicate the lack of knowledge where SFT is intended to teach. A purely self-prior residual treats both cases the same, by reducing the imitation strength and providing no additional guidance over plausible alternatives.

This motivates a residual distribution that remains anchored to the model prior, but also allows external corrective signals. To this end, we construct a teacher-guided $\tilde{\pi}_t$. The goal is to preserve model prior without being constrained by it, enabling supervision from teacher-supported alternatives.

We construct $\tilde{\pi}_t$ through KL-regularized reward shaping:
\begin{align*}
    \tilde{\pi} =
\arg\max_{q\in\Delta}
\left[
\mathbb{E}_{a\sim q}[r(a)]-\tau \mathrm{KL}(q\|\pi_\theta)
\right],
\end{align*}
which stays close to $\pi_\theta$ while the reward $r$ specifies the alternative tokens to be upweighted. Let $\pi_T(\cdot\mid x_t)$ denote a teacher distribution. To incorporate teacher guidance, we define reward using the teacher log-probability,  $r(a)=\lambda\log \pi_T(a)$.

The solution has the form $\tilde{\pi}(a) \propto \pi_\theta(a)\exp(r(a)/\tau)$, and substituting the teacher reward yields
\begin{align*}
    \tilde{\pi}(a)\propto \pi_\theta(a)\pi_T(a)^\eta,
\qquad \eta=\lambda/\tau.
\end{align*}
This results in a teacher-guided $\tilde{\pi}$ close to $\pi_\theta$, while upweighting alternatives favored by the teacher through token-level reward. For easier interpretation, we consider the closely related form
\begin{align*}
\tilde{\pi}_t^{\mathrm{guided}}(a)
\propto
\pi_\theta(a\mid x_t)^{1-\eta}
\pi_T(a\mid x_t)^\eta,
\qquad \eta\in[0,1].
\end{align*}
It interpolates between the student ($\eta \rightarrow 0$) and teacher distribution ($\eta \rightarrow 1$). This parameterization is convenient in practice, since $\eta$ directly controls the intensity of teacher signals.

\paragraph{\Ours{}.}
Combining the probability-proxy for trust estimate $\gamma_t=p_y$ and the teacher-guided residual distribution $\tilde{\pi}=\tilde{\pi}_\text{guided}$ gives the following target:
\begin{align}
Q_t^{\ours}
=
p_y\delta_{y_t}
+
(1-p_y)\tilde{\pi}_t^{\mathrm{guided}}.
\end{align}
The corresponding token-level objective decomposes as
\begin{align}
\mathrm{CE}\!\left(Q_t^{\ours},\pi_\theta\right)
=
p_y\,
\mathrm{CE}\!\left(\delta_{y_t},\pi_\theta\right)
+
(1-p_y)\,
\mathrm{CE}\!\left(\tilde{\pi}_t^{\mathrm{guided}},\pi_\theta\right).
\end{align}

\Ours{} adaptively balances strict imitation and prior preservation. When the observed token $y_t$ is well-supported by the model, the objective approaches standard SFT. When $y_t$ is uncertain, it weakens one-hot fitting and instead assigns a higher weight to the teacher-guided residual branch. In this regime, teacher supervision acts as a \textit{fallback supervision} that (1) avoids overfitting to uncertain labels, and (2) strengthens signals when a desired token is under-supported by student due to low $p_y$.
\section{Experiments}

\subsection{Setup}

For mathematical reasoning, we train on two datasets: \textbf{NuminaMath-CoT-67k}~\cite{numinamath, beyondlog} and \textbf{OpenR1-Math-15k}~\cite{openr1math, luffy}. For broader scientific reasoning, we train on \textbf{m23k}~\cite{m23k}, a high-quality medical reasoning dataset. Our experiments cover across seven diverse models: Qwen2.5 (1.5B \& 7B), Qwen2.5-Math (1.5B \& 7B), Qwen3-1.7B-Base, LlaMA-3.2-3B, LlaMA-3.1-8B.

We compare \Ours{} against the following baselines: (1) \textbf{SFT}, which trains with the standard negative log-likelihood. (2) \textbf{SFT ($p$-loss)}~\cite{beyondlog, dft}, a probability-weighted variant that scales loss by the model probability on the observed token. (3) \textbf{Knowledge Distillation}~\cite{hintondistill}, which applies teacher distribution on the data as supervision. We use the standard form $\mathcal{L}_{\mathrm{Distill}} = c\,\mathrm{CE}(\pi_T,\pi_\theta) + (1-c)\,\mathrm{CE}(\delta_{y_t},\pi_\theta)$, where $c$ is constant and ablated in Table~\ref{tab:ablate}. These baselines correspond to different partial choices in the \(Q\)-target design space: SFT uses the one-hot target, SFT (\(p\)) uses \(\gamma_t=p_{y_t}\), and KD uses the full teacher distribution as a fixed distributional signal. \(\Ours{}\) considers both \((\gamma_t,\tilde{\pi}_t)\) and adaptively balances label imitation and the residual branch based on uncertainty in label. 

For models that involve teacher distribution, we use the corresponding instruction-tuned model as the teacher; for example, Qwen2.5-1.5B uses Qwen2.5-1.5B-Instruct. For the Qwen3 series, we use Qwen3-4B-Instruct-2507 as the teacher. 
The evaluation performance is measured by Average@16 accuracy. Further details of evaluation and training configurations are provided in Appendix~\ref{app:experiment_details}.

\subsection{Main Results}

\begin{figure}[t]
  \centering
  \includegraphics[width=\linewidth]{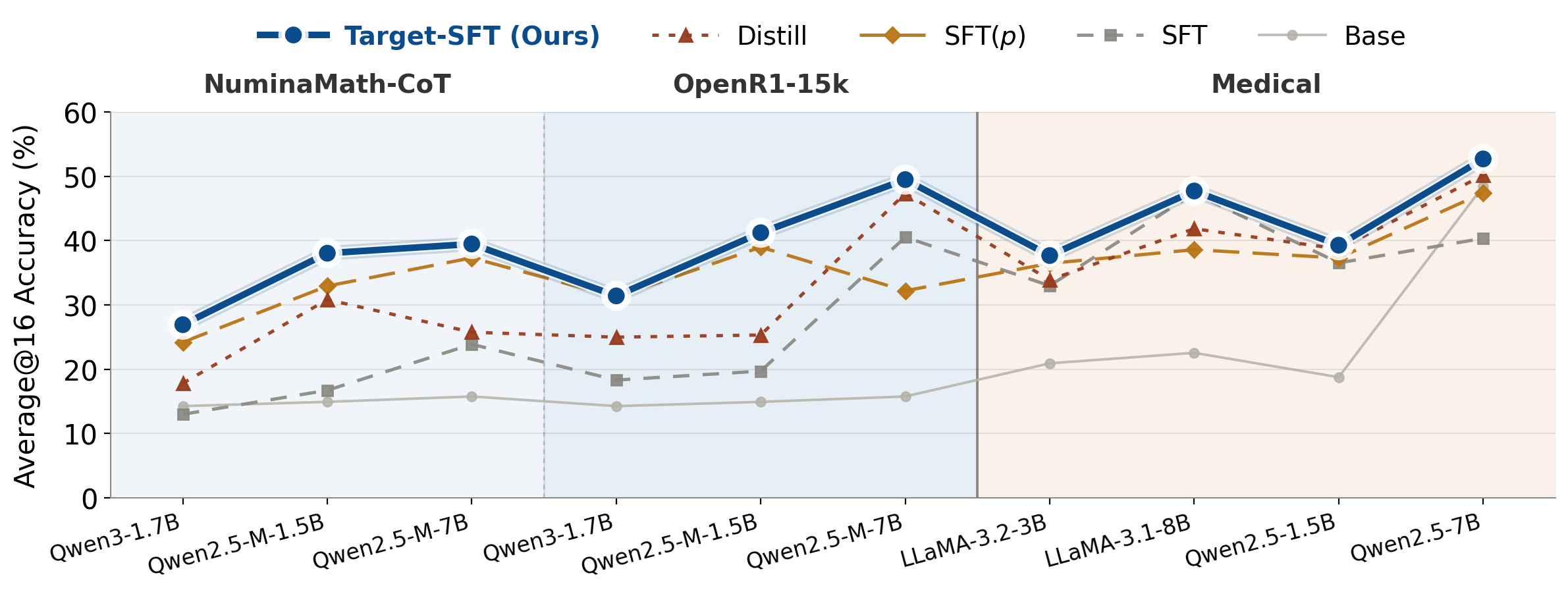}
  \caption{\textbf{Performance Summary.} Average@16 accuracy across all 10 dataset-model settings used.}
  \label{fig:summary}
\end{figure}

Across all evaluations, \Ours{} achieves the highest Average@16 accuracy. Figure~\ref{fig:summary} summarizes the results across tasks. While the baselines each show complementary strengths and weaknesses in different dataset-model settings, \Ours{} consistently gives the best results. This highlights the value of the structure exposed by the $Q$-target framework: effective SFT can arise from carefully designing both the imitation strength and the allocation of remaining probability mass.

\begin{table}[htbp]
  \caption{\textbf{Mathematical Reasoning.} Average@16 accuracy on five standard benchmarks, using models trained on NuminaMath-CoT (top) and OpenR1 (bottom).}
  \label{tab:math_eval}
    \centering
    \fontsize{5.8pt}{6.4pt}\selectfont
    \setlength{\tabcolsep}{8pt}
    \renewcommand{\arraystretch}{0.95}
    \resizebox{\textwidth}{!}{
    \begin{tabular}{lcccccc}
    \toprule
    & \textbf{Minerva Math} & \textbf{Olympiad Bench} & \textbf{AIME24} & \textbf{AMC23} & \textbf{Math500} & \textbf{Avg.} \\
    \midrule
    \multicolumn{7}{c}{\emph{Dataset: NuminaMath-CoT}} \\
    \midrule
    \multicolumn{7}{c}{\textbf{Qwen3-1.7B-Base}} \\
    \midrule
    Base      & 9.87  & 11.35 & 0.62 & 16.25 & 33.81 & 14.26 \\
    SFT       & 10.78 & 8.64  & 0.00 & 10.78 & 34.61 & 12.99 \\
    SFT ($p$) & \underline{18.91} & \underline{17.81} & 1.24 & \underline{27.66} & \underline{53.86} & \underline{24.20} \\
    Distill   & 14.08 & 11.92 & \underline{1.65} & 19.69 & 41.52 & 17.81 \\
    \rowcolor{lightblue}
    \Ours{}   & \textbf{21.44} & \textbf{19.21} & \textbf{3.94} & \textbf{30.78} & \textbf{57.55} & \textbf{26.93} \\
    \midrule
    \multicolumn{7}{c}{\textbf{Qwen2.5-Math-1.5B}} \\
    \midrule
    Base      & 8.23  & 15.20 & 3.75 & 18.12 & 31.52 & 14.92 \\
    SFT       & 12.61 & 12.11 & 0.82 & 16.41 & 42.29 & 16.72 \\
    SFT ($p$) & 25.19 & \underline{27.38} & \underline{7.72} & \underline{38.12} & \underline{65.79} & \underline{32.94} \\
    Distill   & \underline{25.46} & 23.64 & 6.68 & 37.50 & 60.92 & 30.83 \\
    \rowcolor{lightblue}
    \Ours{}   & \textbf{32.20} & \textbf{31.59} & \textbf{8.96} & \textbf{47.03} & \textbf{70.20} & \textbf{38.05} \\
    \midrule
    \multicolumn{7}{c}{\textbf{Qwen2.5-Math-7B}} \\
    \midrule
    Base      & 7.66  & 9.62  & 8.13 & 19.84 & 31.98 & 15.76 \\
    SFT       & 21.33 & 18.77 & 2.70 & 22.81 & 53.55 & 23.88 \\
    SFT ($p$) & \underline{28.48} & \underline{32.78} & \underline{8.56} & \underline{49.38} & \underline{67.93} & \underline{37.33} \\
    Distill   & 23.35 & 18.79 & 5.63 & 30.31 & 51.24 & 25.76 \\
    \rowcolor{lightblue}
    \Ours{}   & \textbf{31.03} & \textbf{34.56} & \textbf{8.96} & \textbf{49.69} & \textbf{72.69} & \textbf{39.49} \\
    \specialrule{1.0pt}{2pt}{2pt}
    \addlinespace[1pt]
    \multicolumn{7}{c}{\emph{Dataset: OpenR1-15k}} \\
    \midrule
    \multicolumn{7}{c}{\textbf{Qwen3-1.7B-Base}} \\
    \midrule
    Base      & 9.87  & 11.35 & 0.62 & 16.25 & 33.81 & 14.26 \\
    SFT       & 14.30 & 12.55 & 1.65 & 20.62 & 41.68 & 18.31 \\
    SFT ($p$) & \underline{25.86} & \underline{23.66} & \textbf{7.29} & \textbf{35.62} & \underline{62.41} & \underline{31.23} \\
    Distill   & 22.43 & 18.42 & 2.90 & 27.81 & 53.70 & 24.99 \\
    \rowcolor{lightblue}
    \Ours{}   & \textbf{27.47} & \textbf{24.78} & \underline{5.41} & \underline{33.75} & \textbf{63.92} & \textbf{31.41} \\
    \midrule
    \multicolumn{7}{c}{\textbf{Qwen2.5-Math-1.5B}} \\
    \midrule
    Base      & 8.23  & 15.20 & 3.75  & 18.12 & 31.52 & 14.92 \\
    SFT       & 14.45 & 15.35 & 1.87  & 24.22 & 43.99 & 19.71 \\
    SFT ($p$) & \underline{31.18} & \underline{33.56} & \underline{11.45} & \underline{47.34} & \underline{70.75} & \underline{39.00} \\
    Distill   & 19.89 & 20.31 & 5.01  & 26.72 & 52.65 & 25.32 \\
    \rowcolor{lightblue}
    \Ours{}   & \textbf{33.09} & \textbf{34.84} & \textbf{13.13} & \textbf{51.72} & \textbf{72.38} & \textbf{41.24} \\
    \midrule
    \multicolumn{7}{c}{\textbf{Qwen2.5-Math-7B}} \\
    \midrule
    Base      & 7.66  & 9.62  & 8.13  & 19.84 & 31.98 & 15.76 \\
    SFT       & 34.41 & 33.26 & 11.88 & 50.16 & 73.29 & 40.55 \\
    SFT ($p$) & 27.33 & 28.28 & 11.24 & 37.19 & 57.25 & 32.20 \\
    Distill   & \underline{42.04} & \underline{39.94} & \underline{13.32} & \textbf{62.34} & \underline{80.26} & \underline{47.36} \\
    \rowcolor{lightblue}
    \Ours{}   & \textbf{43.61} & \textbf{42.42} & \textbf{18.13} & \underline{61.88} & \textbf{80.75} & \textbf{49.50} \\
    \bottomrule
  \end{tabular}
  }
\end{table}

\paragraph{Math.}
Table~\ref{tab:math_eval} reports the results on NuminaMath-CoT and OpenR1-15k. \Ours{} achieves the highest average accuracy across all models on both datasets. In contrast, standard SFT yields only modest gains over the base model in many cases, and even hurts performance for Qwen3-1.7B on NuminaMath-CoT ($14.26 \rightarrow 12.99$). This is consistent with prior observations that rigid one-hot matching can be limiting for mathematical reasoning~\cite{beyondlog, rft, metamath, star}. The relatively strong performance of probability-weighted SFT ($p$-loss) further suggests that reducing imitation strength on uncertain tokens is beneficial. However, \Ours{} improves over the $p$-loss in every case. This supports our claim that choosing $\gamma_t$ alone and defaulting the residual mass to model prior is incomplete. By explicitly designing $\tilde{\pi}_t$, \Ours{} enhances the supervision and achieves stronger performance.

Direct distillation, which uses teacher distribution as the full target, performs relatively weakly on mathematical reasoning. On NuminaMath-CoT, distillation is only slightly better than standard SFT in some cases, such as Qwen3-1.7B-Base ($17.81$ vs. $12.99$) and Qwen2.5-Math-7B ($25.76$ vs. $23.88$), whereas \Ours{} outperforms significantly. This suggests that simply replacing the target with a soft teacher distribution is still suboptimal. In contrast, \Ours{} uses the teacher to shape the residual branch, whose weight $1-\gamma_t$ increases when $y_t$ is under-supported. This adaptive use of teacher signals proves to be more effective than full distillation in these settings.

\paragraph{Medical.}
Table~\ref{tab:medical} shows a different pattern on medical reasoning. The gap between standard SFT and $p$-loss is smaller than in math, and in some cases standard SFT is clearly stronger. For example, on Qwen2.5-1.5B the two methods achieve similar averages, while on LLaMA-3.1-8B standard SFT substantially outperforms $p$-loss ($47.41$ vs. $38.60$). This suggests that for some tasks, stricter imitation is more effective, perhaps because the demonstrations align more closely with the desired answer distribution. In such cases, the $p$-loss that uniformly weakens supervision for all low-$p_y$ tokens is not ideal. Nevertheless, \Ours{} still achieves the best average performance across all models on the medical setting. This shows that the teacher-guided residual branch remains valuable even when probability-based reweighting alone is less effective.
Distillation is also more competitive on medical reasoning than on math, outperforming $p$-loss on Qwen2.5-1.5B and Qwen2.5-7B. However, \Ours{} still outperforms distillation across all models. The mixed performance of distillation again suggests that teacher information is useful, but treating it as a fixed full target is not always the best approach. Under the $Q$-target design, the teacher instead acts as adaptive fallback supervision, which supplements corrective signals for missing knowledge while preserving the model prior.

\begin{table}
  \caption{\textbf{Medical Reasoning.} Average@16 accuracy on medical reasoning benchmarks.}
  \label{tab:medical}
  \centering
  \scriptsize
  \setlength{\tabcolsep}{5pt}
  \renewcommand{\arraystretch}{1.08}
  \resizebox{\textwidth}{!}{
  \begin{tabular}{l*{11}{c}}
    \toprule
    & \textbf{MedMC} & \textbf{MedQA} & \textbf{PubMed} & \textbf{MMLU-P} & \textbf{GPQA} & \textbf{Lancet} & \textbf{MedB (4)} & \textbf{MedB (5)} & \textbf{MedX} & \textbf{NEJM} & \textbf{Avg.} \\
    \midrule
    \multicolumn{12}{c}{\textbf{LLaMA-3.2-3B}} \\
    \midrule
    Base      & 21.13 & 21.76 & 22.0 & 12.18 & 25.64 & 24.51 & 27.92 & 21.43 & 11.11 & 21.56 & 20.92 \\
    SFT       & 34.19 & 38.02 & 57.0 & 25.93 & \underline{30.26} & 36.17 & \underline{36.04} & 28.25 & \underline{11.73} & 32.34 & 32.99 \\
    SFT ($p$) & \textbf{40.43} & \underline{40.93} & \underline{61.4} & \textbf{33.42} & \textbf{34.87} & \underline{43.45} & 34.09 & 25.97 & 9.87 & \textbf{40.30} & \underline{36.47} \\
    Distill   & 37.08 & 38.26 & 55.7 & 28.01 & 27.44 & 39.81 & 35.71 & \textbf{31.82} & 10.01 & 35.16 & 33.90 \\
    \rowcolor{lightblue}
    \Ours{}   & \underline{39.78} & \textbf{44.46} & \textbf{64.0} & \underline{32.70} & 30.00 & \textbf{45.87} & \textbf{37.01} & \underline{31.17} & \textbf{12.63} & \underline{39.30} & \textbf{37.69} \\
    \midrule
    \multicolumn{12}{c}{\textbf{LLaMA-3.1-8B}} \\
    \midrule
    Base      & 22.81 & 29.30 & 21.2 & 19.02 & 29.49 & 22.82 & 29.87 & 20.78 & 10.14 & 20.07 & 22.55 \\
    SFT       & \textbf{50.47} & \underline{56.64} & \underline{74.0} & \underline{48.21} & \underline{37.69} & \underline{53.40} & \textbf{47.73} & \textbf{40.26} & \textbf{14.77} & \textbf{50.91} & \underline{47.41} \\
    SFT ($p$) & 40.69 & 46.90 & 64.3 & 34.33 & 33.33 & 38.35 & 40.58 & 31.82 & 12.22 & 43.45 & 38.60 \\
    Distill   & 45.52 & 51.61 & 63.3 & 42.35 & 35.64 & 47.57 & 38.31 & 35.71 & 13.18 & 45.44 & 41.86 \\
    \rowcolor{lightblue}
    \Ours{}   & \underline{49.32} & \textbf{60.17} & \textbf{74.7} & \textbf{50.42} & \textbf{46.15} & \textbf{55.34} & \underline{41.56} & \underline{38.96} & \underline{13.60} & \underline{47.10} & \textbf{47.73} \\
    \midrule
    \multicolumn{12}{c}{\textbf{Qwen2.5-1.5B}} \\
    \midrule
    Base      & 22.40 & 22.70 & 18.4 & 11.47 & 17.18 & 23.06 & 24.35 & 17.21 & 9.45  & 21.39 & 18.76 \\
    SFT       & 39.35 & 41.16 & \underline{68.5} & 34.07 & 34.36 & 39.32 & 35.39 & \underline{30.19} & 10.35 & 32.67 & 36.54 \\
    SFT ($p$) & 38.92 & 37.55 & 67.6 & 37.79 & 35.64 & \underline{42.72} & 35.71 & \underline{30.19} & 10.49 & 36.48 & 37.31 \\
    Distill   & \textbf{41.02} & \textbf{42.58} & \textbf{68.9} & \underline{37.92} & \underline{38.21} & \textbf{43.20} & \underline{37.01} & 28.57 & \underline{10.70} & \textbf{38.97} & \underline{38.71} \\
    \rowcolor{lightblue}
    \Ours{}   & \underline{40.31} & \underline{41.87} & 68.3 & \textbf{39.67} & \textbf{42.56} & 40.29 & \textbf{38.31} & \textbf{31.49} & \textbf{11.59} & \underline{38.64} & \textbf{39.30} \\
    \midrule
    \multicolumn{12}{c}{\textbf{Qwen2.5-7B}} \\
    \midrule
    Base      & 51.35 & 57.03 & 69.7 & 54.01 & \underline{45.64} & 56.80 & 42.21 & 40.26 & 12.22 & \underline{55.56} & 48.48 \\
    SFT       & 42.24 & 44.07 & 69.3 & 41.43 & 36.67 & 42.96 & 38.64 & 37.01 & 12.01 & 39.30 & 40.36 \\
    SFT ($p$) & 47.65 & 52.79 & \underline{74.5} & 54.27 & 44.62 & 57.04 & 44.81 & 38.96 & 12.97 & 46.60 & 47.42 \\
    Distill   & \underline{52.28} & \underline{58.84} & 72.3 & \underline{58.89} & 40.77 & \underline{58.01} & \underline{50.65} & \underline{41.56} & \underline{13.73} & 54.73 & \underline{50.18} \\
    \rowcolor{lightblue}
    \Ours{}   & \textbf{54.53} & \textbf{62.37} & \textbf{74.9} & \textbf{60.85} & \textbf{48.21} & \textbf{59.95} & \textbf{52.92} & \textbf{42.86} & \textbf{14.01} & \textbf{56.55} & \textbf{52.72} \\
    \bottomrule
  \end{tabular}
  }
\end{table}


\subsection{Ablation Study.}
We ablate the two key design choices in \Ours{}. We include results on varying (1) \textit{the intensity of teacher supervision} in the residual distribution, controlled by $\eta$. This tests the effect of the teacher model and the method's sensitivity to this hyperparameter $eta$. We further vary (2) \textit{the adaptive weighting of the residual branch}, controlled by $1-\gamma_t$. This ablation directly tests whether using an uncertainty-dependent residual weight $1-\gamma_t$ is effective, compared to simply assigning a fixed constant weight $c$ to the $\tilde{\pi}_t$ branch. Additionally, we also tune the hyperparameter $c$ for distillation to explore potential performance gain for the baseline.

We report the full results in Appendix~\ref{app:ablation}. In summary, the ablation studies validate both components of \Ours{}. While varying $\eta \in \{0.2, 0.5, 1.0\}$ gives averages in the range from $34.30$ to $38.05$, they all outperform the baselines and the best distillation setting ($34.30$ vs. $30.83$). In contrast, further tuning of the hyperparameter $c \in \{0.2, 0.5, 0.8, 1.0\}$ for the distillation baseline does not lead to gains beyond the result presented in Table~\ref{tab:math_eval}, but instead fluctuates significantly from $22.81$ to $30.83$. The ablation on residual weight $1-\gamma_t$ also shows the effectiveness of this design, where changing to this to a constant $c$ degrades the performance.

\section{Conclusion}

In this work, we show that supervised fine-tuning is fundamentally a target distribution design. We formalize this view through the $Q$-target framework $Q_t=\gamma_t\delta_{y_t}+(1-\gamma_t)\tilde{\pi}_t$, which exposes two hidden design choices: how strongly to imitate the observed token, and how to allocate residual probability mass over alternatives. This lens unifies many existing SFT variants as implicit choices of $\gamma_t$ or $\tilde{\pi}_t$. Building on this insight, we present \Ours{} and empirically demonstrate its effectiveness across ten reasoning data-model settings. 
Overall, our results offer a novel and more complete perspective into SFT, and open a broader design space for future SFT methods.

\newpage
\bibliographystyle{unsrt} 
\bibliography{Reference}


\newpage
\appendix

\section{Experiment Details}
\label{app:experiment_details}

\paragraph{Training Configurations.} All SFT experiments are conducted using \texttt{verl}, largely following the work by Li et al.~\cite{beyondlog}. The optimizer used is AdamW, with the learning rate of $5 \times 10^{-5}$ for all models. We use cosine decay scheduling with a warm-up ratio of 0.1. We use the maximum sequence length of 3072 for both mathematical reasoning datasets. The global train batch size is 256, with gradient accumulation. The experiments are conducted on A6000 and H200 GPUs. All models are trained for 1 epoch. Table~\ref{tab:training_config} summarizes all training and evaluation-related configurations.

\paragraph{Teacher Models.} For methods that involve teacher distribution, we use the corresponding instruction-tuned model as the teacher for the base model; for example, Qwen2.5-1.5B uses Qwen2.5-1.5B-Instruct. For Qwen3-1.7B-Base, we use Qwen3-4B-Instruct-2507 as the teacher. For \Ours{}, we choose the teacher signal intensity from $\eta\in \{0.2, 0.5, 1.0\}$. To ensure consistency across methods, we cache all teacher logits before training and use the same cache for both distillation and \Ours{} experiments. For memory efficiency, the cached teacher distribution is truncated to the top-64 tokens in the vocabulary with the highest probabilities.

\paragraph{Datasets.} For NuminaMath-CoT which originally contains $859$k chain-of-through problems, we following Li et al.~\cite{beyondlog} and use the $67$k subset organized in their work. For the OpenR1 training dataset, we sample $15$k from OpenR1-Math-46k-8192~\cite{luffy}, which contains verified traces generated by DeepSeek-R1 for problems collected from NuminaMath 1.5~\cite{numinamath}. We collect the responses with sequence length shorter than $3072$ to standardize with the training configuration for NuminaMath-CoT. For scientific reasoning, we use m23k~\cite{m23k}, a $23$k high-quality medical reasoning dataset.

\paragraph{Evaluation.} For mathematical reasoning, evaluation covers five representative benchmarks: Minerva Math, Olympiad Bench, AIME24, AMC23, and Math500~\cite{minerva, olympiad, aime24, amc23, math500}. For models trained on m23k, we evaluate on the same benchmarks following Li et al.~\cite{beyondlog}, which include MedMCQA~\cite{medmcqa}, MedQA-USMLE~\cite{MedQA-USMLE}, PubMedQA~\cite{PubMedQA}, MMLU-Pro~\cite{MMLU-Pro},  GPQA (Medical)~\cite{GPQA}, Lancet \&
NEJM~\cite{m23k}, MedBullets~\cite{MedBullets}, and MedXpertQA~\cite{MedXpertQA}. Evaluation follows the same protocol as Li et al.~\cite{beyondlog} and Huang et al~\cite{m23k}. Inference uses a maximum generation length of $4096$ tokens with the decoding temperature 1.0. For distillation and \Ours{} on Qwen2.5-Math-7B trained from the longer-sequence OpenR1-15k dataset, we use the decoding settings with temperature 0.7 and top-p 0.8 as recommended for its teacher model Qwen2.5-Math-7B-Instruct. The reported results are averaged over 16 generations for every prompt.


\begin{table}[htbp]
\centering
\small
\setlength{\tabcolsep}{5pt}
\renewcommand{\arraystretch}{1.15}
\caption{\textbf{Configuration summary.} All experiments use the same setup unless otherwise specified.}
\label{tab:training_config}
\begin{tabular}{lll}
\toprule
\textbf{Section} & \textbf{Item} & \textbf{Details} \\
\midrule
Dataset 
& NuminaMath-CoT 
& $67$k subset from Li et al.~\cite{beyondlog} \\
& OpenR1 
& $15$k samples with length $<3072$ from OpenR1-Math-46k-8192~\cite{luffy} \\
& Medical reasoning 
& m23k~\cite{m23k}, a $23$k high-quality medical reasoning dataset \\
\midrule
Model
& Math
& Qwen3-1.7B-Base, Qwen2.5-Math-1.5B, Qwen2.5-Math-7B \\
& Medical
& LLaMA-3.2-3B, LLaMA-3.1-8B, Qwen2.5-1.5B, Qwen2.5-7B \\
\midrule
Train 
& Framework 
& \texttt{verl} \\
& Optimizer 
& AdamW \\
& Learning rate 
& $5 \times 10^{-5}$ \\
& Schedule 
& Cosine decay with warm-up ratio $0.1$ \\
& Max sequence length 
& $3072$ \\
& Global batch size 
& $256$, with gradient accumulation \\
& Epochs 
& $1$ \\
\midrule
Eval 
& Max generation length 
& $4096$ \\
& Metric 
& Average@16 \\
\bottomrule
\end{tabular}
\end{table}

\newpage
\section{Proofs}
\subsection{Proof of Proposition~\ref{prop:q_decomp}}
\label{app:proof_loss_decomp}
For a target distribution $Q_t$, the token-level cross-entropy loss is
\begin{align*}
\text{CE}\left(Q_t, \pi_\theta(\cdot \mid x_t)\right)
=
-\sum_{k \in \mathcal{V}} Q_t(k)\log \pi_\theta(k \mid x_t).
\end{align*}
By definition, the Q-target is
\begin{align*}
Q_t = \gamma_t \delta_{y_t} + (1-\gamma_t)\tilde{\pi}_t.
\end{align*}

Substituting this into the cross-entropy gives
\begin{align}
\text{CE}\left(Q_t, \pi_\theta(\cdot \mid x_t)\right)
&=
-\sum_{k \in \mathcal{V}}
\left[
\gamma_t \delta_{y_t}(k)
+
(1-\gamma_t)\tilde{\pi}_t(k)
\right]
\log \pi_\theta(k \mid x_t) \notag
\\
&=
-\gamma_t \sum_{k \in \mathcal{V}}
\delta_{y_t}(k)\log \pi_\theta(k \mid x_t)
-
(1-\gamma_t)\sum_{k \in \mathcal{V}}
\tilde{\pi}_t(k)\log \pi_\theta(k \mid x_t) \notag
\\
&=
\gamma_t \text{CE}\left(\delta_{y_t}, \pi_\theta(\cdot \mid x_t)\right)
+
(1-\gamma_t)\text{CE}\left(\tilde{\pi}_t, \pi_\theta(\cdot \mid x_t)\right).
\end{align}

Therefore, training toward $Q_t$ decomposes the token-level supervision into two components: label imitation controlled by $\gamma_t$ and residual distribution matching of $\tilde{\pi}_t$.

\subsection{Proof of Corollary~\ref{cor:token_variant}}
\label{app:proof_token_variant}
Given the token-weighted loss with detached importance weighting $w_t$
\begin{align*}
    \mathcal{L}_t^{w}
    =
    -w_t \log \pi_\theta(y_t\mid x_t),
\end{align*}
we consider the choice of $(\gamma_t, \tilde{\pi}_t)$
\begin{align*}
\left(
\gamma_t=w_t,\,
\tilde{\pi}_t=\mathrm{sg}\!\left[\pi_\theta(\cdot\mid x_t)\right]
\right)
\quad\Longrightarrow\quad
Q_t = w_t\delta_{y_t} + (1-w_t)\mathrm{sg}\!\left[\pi_\theta(\cdot\mid x_t)\right].
\end{align*}

By Proposition~\ref{prop:q_decomp}, the token-level loss for this $Q_t$ decomposes as
\begin{align*}
    \mathrm{CE}\!\left(Q_t,\pi_\theta\right)
    &=
    w_t
    \mathrm{CE}\!\left(\delta_{y_t},\pi_\theta\right) +
    (1-w_t)
    \mathrm{CE}\!\left(
    \mathrm{sg}\!\left[\pi_\theta\right],
    \pi_\theta
    \right).
\end{align*}

The first term gives the token-weighted SFT gradient:
\begin{align*}
    \nabla_\theta
    w_t
    \mathrm{CE}\!\left(\delta_{y_t},\pi_\theta\right)
    =
    -w_t
    \nabla_\theta
    \log \pi_\theta(y_t\mid x_t).
\end{align*}
For the second term, let $p_t=\pi_\theta$ and $\bar p_t=\mathrm{sg}[p_t]$. Since the logit gradient of $\mathrm{CE}(\bar p_t,p_t)$ is $p_t-\bar p_t$, and $\bar p_t$ is a detached copy of $p_t$, we have
\begin{align*}
    \nabla_\theta
    \mathrm{CE}\!\left(
    \mathrm{sg}\!\left[\pi_\theta\right],
    \pi_\theta
    \right)
    =
    0.
\end{align*}
Therefore,
\begin{align}
    \nabla_\theta
    \mathrm{CE}\!\left(Q_t,\pi_\theta\right)
    =
    -w_t\nabla_\theta\log \pi_\theta(y_t\mid x_t)
    =
    \nabla_\theta \mathcal{L}_t^{w}.
\end{align}
This shows that token-weighted SFT is equivalent at the gradient level to $Q$-target training with $\gamma_t=w_t$ and a self-matching residual branch $\tilde{\pi}_t=\mathrm{sg}\!\left[\pi_\theta\right]$.

\newpage
\subsection{Proof of Corollary~\ref{cor:dist_variant}}
\label{app:proof_dist_variant}
Given the distributional objective
\begin{align*}
    \mathcal{L}_t
    =
    -a_t \log \pi_\theta(y_t\mid x_t)
    +
    b_t\,\mathrm{CE}(q_t,\pi_\theta),
    \qquad a_t,b_t\ge 0,
\end{align*}
rewrite the label imitation term as cross-entropy and obtain:
\begin{align*}
    \mathcal{L}_t
    =
    a_t\,\mathrm{CE}(\delta_{y_t},\pi_\theta)
    +
    b_t\,\mathrm{CE}(q_t,\pi_\theta).
\end{align*}
Let $s_t=a_t+b_t$. For $s_t>0$, this can be normalized as
\begin{align*}
    \mathcal{L}_t
    =
    s_t
    \left[
    \frac{a_t}{s_t}\mathrm{CE}(\delta_{y_t},\pi_\theta)
    +
    \frac{b_t}{s_t}\mathrm{CE}(q_t,\pi_\theta)
    \right].
\end{align*}
The constant $s_t$ only rescales the token-level gradient globally, which can be absorbed into the effective learning rate or token weight. Define
\begin{align*}
    \gamma_t=\frac{a_t}{a_t+b_t},
    \qquad
    \tilde{\pi}_t=q_t.
\end{align*}
By linearity of cross-entropy in its first argument,
\begin{align}
    &\gamma_t\mathrm{CE}(\delta_{y_t},\pi_\theta)
    +
    (1-\gamma_t)\mathrm{CE}(\tilde{\pi}_t,\pi_\theta)
    =
    \mathrm{CE}
    \left(
    \gamma_t\delta_{y_t}
    +
    (1-\gamma_t)\tilde{\pi}_t,
    \pi_\theta
    \right).
\end{align}
Therefore, up to the overall scale $a_t+b_t$, the objective is equivalent to Q-target training with
\begin{align*}
    Q_t
    =
    \gamma_t\delta_{y_t}
    +
    (1-\gamma_t)\tilde{\pi}_t.
\end{align*}
This proves the claim of Corollary~\ref{cor:dist_variant}.

Some distributional methods fall outside the nonnegative residual-mixture form in Eq.~\eqref{eq:distributional_objective}. For example, GEM~\cite{gem} discourages model from concentrating high probability on the strongest non-label alternatives. This can be expressed schematically as a repulsive branch $\tilde{\pi}^-_{t,\tau}$ as follows
\begin{align*}
    \tilde{\pi}_{t,\tau}^{-}(k)
=
\frac{\pi_\theta(k\mid x_t)^{1/\tau}}
{\sum_{v\in\mathcal V}\pi_\theta(v\mid x_t)^{1/\tau}},
\qquad \tau<1.
\end{align*}
This represents a $\tau$-sharpened model distribution, which places mass on high-probability tokens. And its objective is conceptually written as
\begin{align*}
    \mathcal{L}_t
=
\mathrm{CE}(\delta_{y_t},\pi_\theta) - \lambda\,\mathrm{CE}(\tilde{\pi}_{t,\tau}^{-},\pi_\theta),
\end{align*}
shaping the residual branch repulsively against collapsing onto a small set of tokens, thereby aligning with GEM's goal of preserving diversity in SFT.
 
This can be viewed as a signed residual extension of the Q-target framework, with $\tilde{\pi}_t=(\tilde{\pi}_t^+, \tilde{\pi}_t^-)$. This defines a \textit{desired and undesired} target alternative, where $Q_t =\delta_{y_t}+\tilde{\pi}_t^+ + \tilde{\pi}_t^-$. Instead of matching a positive residual distribution, GEM specifies an undesired residual direction to preserve diversity. More generally under this view, distributional variants can shape the residual branch to emphasize or suppress particular subsets of tokens.

\newpage
\section{Unifying Framework}
\label{app:unify}

In this section, we provide concrete connections between SFT variants discussed in Section~\ref{sec:related} and our $Q$-target perspective. Table~\ref{tab:sft_variants} shows the loss formulation and core motivation for each variant. Following the formulation $Q_t=\gamma_t\delta_{y_t}+(1-\gamma_t)\tilde{\pi}_t$, these methods can be interpreted through their choices of $\gamma_t$ and $\tilde{\pi}_t$, as summarized in Table~\ref{tab:unify}. 

\begin{table}[h]
\centering
\footnotesize
\setlength{\tabcolsep}{3pt}
\renewcommand{\arraystretch}{1.28}
\caption{\textbf{Details of SFT Variants.} Each method is presented as a token-level objective \(\ell_t\), given the prefix \(x_t\) and observed token \(y_t\). We denote \(p_t=\pi_\theta(y_t\mid x_t)\), \(p_\theta(v)=\pi_\theta(v\mid x_t)\), \(\mathrm{sg}[\cdot]\) as stop-gradient, \(\pi_T,\pi_S\) as teacher and student distribution, and \(\mathcal V\) as the vocabulary.}
\label{tab:sft_variants}
\newcolumntype{M}{>{\centering\arraybackslash}m{0.52\textwidth}}
\newcolumntype{Y}{>{\raggedright\arraybackslash}p{0.26\textwidth}}
\begin{tabularx}{\textwidth}{p{0.16\textwidth} M Y}
\toprule
\textbf{SFT Variant} & \textbf{Token-level Objective} & \textbf{Motivation} \\
\midrule

Standard SFT
&
\(\ell^{\mathrm{SFT}}_t=-\log p_t\)
&
Maximize likelihood of every observed token \\

\bottomrule
\multicolumn{3}{c}{\rule{0pt}{2.2ex}\textit{Token-Level Variants}} \\
\bottomrule
\\[0.1pt]

DFT~\cite{dft}
&
\(\ell^{\mathrm{DFT}}_t=-\mathrm{sg}[p_t]\log p_t\)
&
Use weighting to connect SFT with an RL-style objective \\
\lightrule

Beyond-log~\cite{beyondlog}
&
\(\begin{gathered}
\ell^f_t=f(p_t),\quad
\ell^\alpha_t=\frac{1-p_t^\alpha}{\alpha}
\end{gathered}\)
&
Use probability-dependent objectives to balance learning across model capacities \\

\lightrule

ProFiT~\cite{profit}
&
\(\begin{gathered}
m_t=\mathbf{1}[\mathrm{sg}(p_t)>\tau],\\
\ell^{\mathrm{ProFiT}}_t=-m_t\log p_t
\end{gathered}\)
&
Use probability to identify and train on core tokens \\

\lightrule

EAFT~\cite{eaft}
&
\(\begin{gathered}
\widetilde H_t
=
\operatorname{sg}\bigg[\frac{
H(\pi_{\theta,t}^{(k)})
}{\log k}\bigg],\\
\qquad
\ell_t^{\mathrm{EAFT}}
= -\widetilde H_t\log p_t .
\end{gathered}\)
&
Use entropy to weight uncertain or knowledge-conflicting tokens \\

\lightrule

iw-SFT~\cite{iwsft}
&
\(\begin{gathered}
w(\tau)=\frac{q(\tau)}{\pi_{\mathrm{ref}}(\tau)}, \text{($w$ trajectory-level)}\\
\ell^{\mathrm{iw}}_t=-w(\tau)\log p_t
\end{gathered}\)
&
Use an auxiliary distribution to assign trajectory-level weights \\

\lightrule

CFT~\cite{cft}
&
\(\begin{gathered}
c_t
=
\mathbf{1}\!\left[
\forall \tilde y_t\in \mathcal A_t,\ 
\mathrm{Correct}
\bigl(
y_{<t},\tilde y_t,y_{>t}
\bigr)=0
\right],\\
\ell^{\mathrm{CFT}}_t=-c_t\log p_t
\end{gathered}\)
&
Update only causally critical / irreplaceable tokens \\

\\[0.1pt]
\bottomrule
\multicolumn{3}{c}{\rule{0pt}{2.2ex}\textit{Distributional-Level Variants}} \\
\bottomrule
\\[0.1pt]

Label Smooth~\cite{sftlabelsmooth}
&
\(\ell^{\mathrm{LS}}_t
=-\big[(1-\lambda)\log p_t+
\frac{\lambda}{|\mathcal V|}\sum_{v\in\mathcal V}\log p_{\theta,t}(v)\big]\)
&
Regularize overconfident predictions for better calibration\\

\lightrule

SFT + KL~\cite{rlsrazor}
&
\(\ell^{\mathrm{KL}}_t
=-\log p_t+\lambda\,
\mathrm{KL}\!\left(
\pi_{\mathrm{ref}}(\cdot\mid x_t)
\middle\|
\pi_\theta(\cdot\mid x_t)
\right)\)
&
Constrain updates with a reference model to limit drift \\
\lightrule

ASFT~\cite{asft}
&
\(\ell^{\mathrm{ASFT}}_t
=\ell^{\mathrm{DFT}}_t+
\lambda \mathrm{KL}\!\left(
\pi_{\mathrm{base}}(\cdot\mid x_t)
\middle\|
\pi_\theta(\cdot\mid x_t)
\right)\)
&
Constrain updates in DFT to prevent distributional drift \\

\lightrule

Proximal SFT~\cite{proximalsft}
&
\(\begin{gathered}
r_t=\frac{p_t}{\pi_{\mathrm{old}}(y_t\mid x_t)},\\
\ell^{\mathrm{PSFT}}_t
=-\min(r_t,\operatorname{clip}(r_t,1-\epsilon,1+\epsilon))
\end{gathered}\)
&
Clip ratio to enforce updates within a trust region \\

\lightrule

GEM~\cite{gem}
&
\(\begin{gathered}
q_t(v)
=
\frac{
\operatorname{sg}[\pi_{\theta,t}(v)]^{1/\beta}
}{
\sum_{u\in\mathcal V}
\operatorname{sg}[\pi_{\theta,t}(u)]^{1/\beta}
},\\
\ell_t^{\mathrm{GEM}}
=
\mathrm{CE}(\delta_{y_t},\pi_{\theta,t})
-
\mathrm{CE}(q_t, \pi_{\theta,t})
\end{gathered}\)
&
Control probability transfer from alternatives to observed token to preserve diversity \\

\lightrule

Knowledge \quad \quad Distillation~\cite{hintondistill}
&
\(\ell^{\mathrm{KD}}_t
=-\sum_{v\in\mathcal V}\pi_T(v\mid x_t)\log \pi_S(v\mid x_t)\)
&
Use the teacher logit distribution as a soft target \\

\lightrule

Distillation \quad \quad(Hybrid)~\cite{hintondistill}
&
\(\begin{gathered}
\ell^{\mathrm{KD\text{-}H}}_t
=(1-\lambda)[-\log \pi_S(y_t\mid x_t)]\\
+\lambda D_{\mathrm{KL}}(\pi_T(\cdot\mid x_t)\| \pi_S(\cdot\mid x_t))
\end{gathered}\)
&
Use hard-label and enrich it with teacher logit distribution \\

\bottomrule
\end{tabularx}
\end{table}


We organize existing methods into three broad categories. \textbf{Label-trust} variants modify the imitation strength on $y_t$, corresponding to different choices of $\gamma_t$. \textbf{Residual-distribution} variants primarily specify where the non-label probability mass should go. In these cases, the corresponding $\gamma_t$ is often obtained by normalizing the relative weights of the hard-label and residual branches, with any overall scale absorbed into the effective learning rate. Finally, \textbf{data-level} methods do not directly alter $\gamma_t$ or $\tilde{\pi}_t$ for a fixed demonstration. Instead, they reshape the empirical target distribution by selecting, filtering, or rewriting the training trajectory $\hat y$, after which standard one-hot SFT is applied. We therefore include them as indirect instances of target distribution design at the dataset level.

In particular, GEM~\cite{gem} can also be illustrated with a signed extension of the $Q$-framework, where the residual component contains both positive and negative branches, $\tilde{\pi}_t=(\tilde{\pi}_t^+, \tilde{\pi}_t^-)$. This defines a \textit{desired and undesired} target alternative, where $Q_t =\delta_{y_t}+\tilde{\pi}_t^+ + \tilde{\pi}_t^-$. And GEM utilizes $\tilde{\pi}_t^-$ as a repulsive branch to discourage probability collapse onto a small set of high-probability tokens, hence preserving diversity. Proximal SFT~\cite{proximalsft} is included as a schematic connection, since its clipping-based trust-region objective is not an explicit residual target but implicitly constrains updates. 

Throughout the table, $(\gamma_t, \tilde{\pi}_t)$ are treated as stop-gradient unless otherwise specified. Some methods are originally defined at the sequence or trajectory level~\cite{iwsft, cft}, we present their token-level decompositions or analogues to make the connection to $Q_t$ explicit.

\begin{table}[htbp]
\centering
\footnotesize
\setlength{\tabcolsep}{5pt}
\renewcommand{\arraystretch}{1.8}
\caption{\textbf{SFT variants under the $Q$-target view.}
Each method can be interpreted through choices of label trust $\gamma_t$
and residual distribution $\tilde{\pi}_t$ in
$Q_t=\gamma_t\delta_{y_t}+(1-\gamma_t)\tilde{\pi}_t$. This table provides illustrative examples rather than an exhaustive coverage of methods in each category.}
\label{tab:unify}
\resizebox{\textwidth}{!}{
\begin{tabular}{llll}
\toprule
\textbf{Variant} 
& \textbf{Category} 
& \textbf{Choice of $\gamma_t$} 
& \textbf{Choice of $\tilde{\pi}_t$} \\
\midrule

Standard SFT
& One-hot imitation
& $1$
& -- \\
\midrule

DFT~\cite{dft}
& Label Trust
& $p_t$
& $\pi_{\theta,t}$ \\

Beyond-log~\cite{beyondlog}
& Label Trust
& $p_t^\alpha$
& $\pi_{\theta,t}$ \\

ProFiT~\cite{profit}
& Label Trust
& $m_t=\mathbf{1}\{p_t>\tau\}$
& $\pi_{\theta,t}$ \\

EAFT~\cite{eaft}
& Label Trust
& $\widetilde H_t=\frac{H(\pi_{\theta,t}^{(k)})}{\log k}$
& $\pi_{\theta,t}$ \\

iw-SFT~\cite{iwsft}
& Label Trust
& $w(\tau)=\frac{q(\tau)}{\pi_{\mathrm{ref}}(\tau)}, \text{(trajectory-level)}$
& $\pi_{\theta,t}$ \\

CFT~\cite{cft}
& Label Trust
& $c_t=\mathbf{1}\{y_t\text{ counterfactual critical}\}$
& $\pi_{\theta,t}$ \\

Label Smoothing~\cite{sftlabelsmooth}
& Residual Distribution
& $1-\lambda$
& $\mathrm{Unif}(\mathcal V)$ \\

SFT + KL~\cite{rlsrazor}
& Residual Distribution
& $\frac{1}{1+\lambda}$ 
& $\pi_{\mathrm{ref}}(\cdot\mid x_t)$ \\

ASFT~\cite{asft}
& Residual Distribution
& $\frac{p_t}{p_t+\lambda}$
& $\pi_{\mathrm{base}}(\cdot\mid x_t)$ \\

Proximal SFT~\cite{proximalsft}
& Residual Distribution
& clipping-dependent
& $\pi_{\mathrm{old}}(\cdot\mid x_t)$ \\

GEM~\cite{gem}
& Residual Distribution
& $\gamma_t^{y}=1,\ \gamma_t^{-}=1$ & $\tilde{\pi}^{+}_{t}=\pi_{\theta,t}$, $\tilde{\pi}^{-}_{t} =\pi^{(\beta)}_{\theta,t}$
\\

Knowledge Distillation~\cite{hintondistill}
& Residual Distribution
& $0$
& $\pi_T(\cdot\mid x_t)$ \\

Distillation (Hybrid)~\cite{hintondistill}
& Residual Distribution
& $1-\lambda$
& $\pi_T(\cdot\mid x_t)$ \\

RFT~\cite{rft}
& Data-Level 
& -- 
& $\delta_{\hat{y}},\hat{y}\sim\pi_\text{gen, correct}(\cdot |x)$ \\

STaR~\cite{star}
& Data-Level 
& -- 
& $\delta_{\hat{y}},\hat{y}\sim\pi_\text{gen, correct}(\cdot |x)$ \\

GRAPE~\cite{grape}
& Data-Level 
& -- 
& $\delta_{\hat{y}},\hat{y} = \arg\max_{y^{(i)}} \pi_{\theta_0}(y^{(i)}\mid x)$ \\

Self-distillation~\cite{selfdistillft}
& Data-Level
& -- 
& $\delta_{\hat{y}},\hat{y} = \mathrm{Rewrite}_{\pi_{\theta_0}}(x,y)$ \\

\midrule

\Ours{}
& Label Trust + Residual
& $p_t$
& $\tilde{\pi}^{\mathrm{guided}}_t \propto \pi_{\theta,t}^{1-\eta}\pi_T(\cdot\mid x_t)^\eta$ \\

\bottomrule
\end{tabular}
}
\end{table}

\newpage
\section{From Any Loss to $Q_t$}
\label{app:loss-to-q}

Section~\ref{sec:any_loss_to_Q} introduces the derivation of $Q$-target from any differentiable token-level SFT loss. We now provide two examples and visualize their effects below.

\paragraph{Example 1: Standard SFT.}
Standard SFT minimizes the negative log-likelihood of the observed token $ \mathcal{L}_{\mathrm{SFT}}(z) = -\log p_y,$ where $p_y$ is the model probability assigned to $y$. The logit gradient is
\begin{align}
    g_j
=
\frac{\partial \mathcal{L}_{\mathrm{SFT}}}{\partial z_j}
=
p_j-\mathbf{1}\{j=y\}.
\end{align}
Substituting this into Eq.~\eqref{eq:induced_Q}, we obtain the induced target
\begin{align}
    Q_{\mathrm{SFT}}(j)
=
p_j-g_j
=
\begin{cases}
1, & j=y,\\
0, & j\neq y.
\end{cases}
\end{align}
This recovers exactly the one-hot target distribution $\delta_y$, assigning all target mass to the observed token and zero mass to alternatives.

\paragraph{Example 2: Probability-Weighted SFT.}
Consider the detached probability-weighted loss
\begin{align}
    \mathcal{L}_{\text{p-loss}}(z)
=
-\mathrm{sg}[p_y]\log p_y,
\end{align}
This loss has the logit gradient
\begin{align}
    g_j
=
\mathrm{sg}[p_y](p_j-\mathbf{1}\{j=y\}).
\end{align}
Substituting this gradient into Eq.~\eqref{eq:induced_Q}, and using the scalar $\mathrm{sg}[p_y]=p_y$ derives the induced target
\begin{align}
    Q_{\text{p-loss}}(j)
=
\begin{cases}
2p_y-p_y^2, & j=y,\\
(1-p_y)p_j, & j\neq y.
\end{cases}
\end{align}
This closed form reveals the mechanism of probability-weighted SFT. When the model assigns high probability to the observed token ($p_y\rightarrow 1$), the induced target approaches the one-hot SFT target. When the model is uncertain ($p_y\rightarrow 0$), the induced target relaxes toward the model’s own distribution $Q_j\rightarrow p_j$. Thus, low-confidence tokens receive weaker imitation updates, and the residual probability mass defaults to the student prior. In the $Q$-target notation, this corresponds to choosing $\gamma=p_y$ and $\tilde{\pi}=\mathrm{sg}[\pi_\theta]$, thereby preserving the model prior when evidence for strict imitation on $y$ is weak. 

\begin{figure}[htbp]
  \centering
  \includegraphics[width=0.8\linewidth]{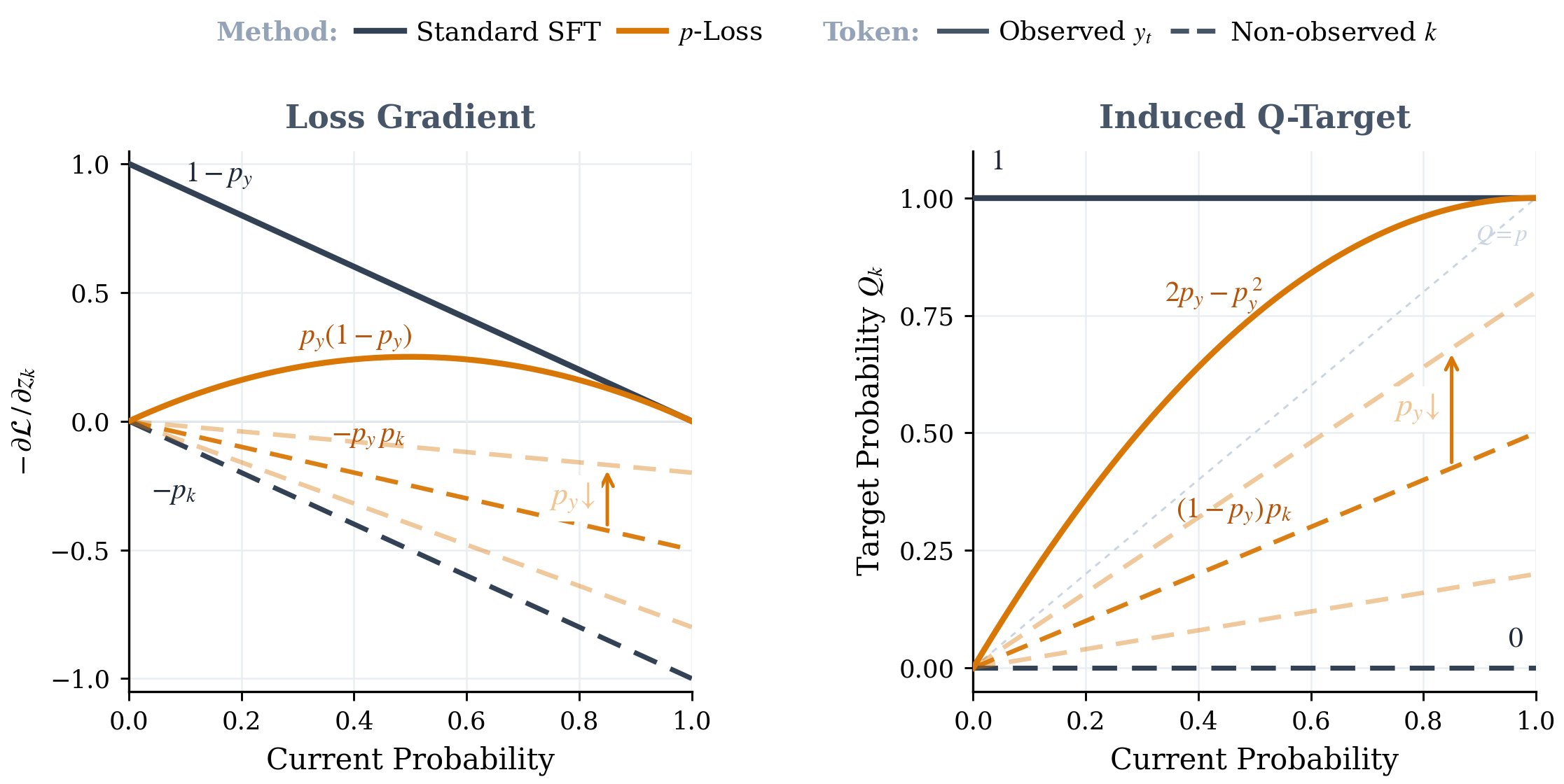}
  \caption{\textbf{Visualization of Loss.} Standard SFT's gradient pulls toward $y_t$ and suppresses all $k$ with fixed strength, corresponding to an induced target $\delta_{y_t}$. For $p$-loss, the gradient scales with $p_y$ (slope for Non-observed token $k$ depends on $p_y$). Therefore, its gradient is near-zero when $p_y\approx 0$, and the target probability is the same as current probability ($Q=p$); the induced target approaches $\delta_{y_t}$ only when $p_y \rightarrow 1$, where the model is certain. An interactive plot is available on our \href{\projecturl}{project page}.}
  \label{fig:loss_visual}
\end{figure}

\newpage
\section{Ablation Study}
\label{app:ablation}

We ablate on the two key design choices in \Ours{}: (1) \textbf{intensity of teacher supervision} in the residual distribution, controlled by $\eta$, and (2) \textbf{adaptive weighting of the residual branch}, controlled by $1-\gamma_t$. The first ablation tests the effect of teacher model and the method's sensitivity to this hyperparameter $\eta$. The second ablation tests whether using an uncertainty-dependent residual weight $1-\gamma_t$ is useful, compared to simply assigning a fixed constant weight $c$ to the $\tilde{\pi}_t$ branch.

Additionally, we ablate on the hyperparameter $c=\{0.2,0.5,0.8,1.0\}$ in the knowledge distillation baseline $\bigl(\mathcal{L}_{\mathrm{Distill}} = c\,\mathrm{CE}(\pi_T,\pi_\theta) + (1-c)\,\mathrm{CE}(\delta_{y_t},\pi_\theta)\bigr)$. This presents further results on baseline performance beyond the default choice of $c=0.8$ used in the main text.

\begin{table}[htbp]
  \caption{\textbf{Ablation Study.} Average@16 accuracy using Qwen2.5-Math-1.5B trained on NuminaMath (top) and OpenR1 (bottom) to ablate on two key designs of the method.}
  \label{tab:ablate}
  \centering
  \scriptsize
  \setlength{\tabcolsep}{7pt}
  \renewcommand{\arraystretch}{1.08}
  \resizebox{\textwidth}{!}{
  \begin{tabular}{lcccccc}
    \toprule
    & \textbf{Minerva Math} & \textbf{Olympiad Bench} & \textbf{AIME24} & \textbf{AMC23} & \textbf{Math500} & \textbf{Avg.} \\
    \midrule
    \multicolumn{7}{c}{\emph{Intensity of teacher signal $\eta$ ($\eta \uparrow =$ dominant supervision)}} \\
    \midrule
    Distill ($c=0.2$)     & 13.99 & 13.08 & 1.45 & 17.34 & 44.32 & 18.33 \\
    Distill ($c=0.5$)     & 19.13 & 18.94 & 4.16 & 26.56 & 53.81 & 24.92 \\
    Distill ($c=0.8$)     & 25.46 & 23.64 & 6.68 & 37.50 & 60.92 & 30.83 \\
    Distill ($c=1.0$)     & 18.07 & 18.49 & 4.38 & 28.28 & 45.22 & 22.81 \\
    \rowcolor{lightgreen}
    \Ours{} ($\eta=0.2$)  & 27.41 & 29.48 & 9.17 & 44.06 & 65.96 & 35.16 \\
    \rowcolor{lightgreen}
    \Ours{} ($\eta=0.5$)  & 32.20 & 31.59 & 8.96 & 47.03 & 70.20 & 38.05 \\
    \rowcolor{lightgreen}
    \Ours{} ($\eta=1.0$)  & 28.24 & 28.13 & 7.92 & 43.12 & 63.85 & 34.30 \\
    \midrule
    \multicolumn{7}{c}{\emph{Constant branch $c$, instead of residual $1-\gamma_t$}} \\
    \midrule
    $c=0.2$               & 22.09 & 23.23 & 5.00 & 33.44 & 58.24 & 28.48 \\
    $c=0.5$               & 29.54 & 33.19 & 11.67 & 50.47 & 70.71 & 39.10 \\
    $c=1.0$               & 22.27 & 27.26 & 8.14 & 41.72 & 60.32 & 31.82 \\
    \rowcolor{lightgreen}
    \Ours{}               & 33.09 & 34.84 & 13.13 & 51.72 & 72.38 & 41.24 \\
    \bottomrule
  \end{tabular}
  }
\end{table}

The results in Table~\ref{tab:ablate} show that \Ours{} still achieves the highest results among all baselines. On the other hand, knowledge distillation in this setting is highly sensitive to the mixture weight $c$. In particular, it requires intricate balancing between teacher and hard-label supervision, where $c=0.5$ and full distillation with $c=1.0$ both degrade significantly from $c=0.8$. And further tuning of this hyperparameter did not provide additional benefits, where the highest-performing distillation baseline still remains far below \Ours{}. This suggests that although the teacher distribution provides useful supervision, using it as a full or fixed soft target is still suboptimal.

The second ablation further confirms the importance of adaptive residual weighting. Replacing the uncertainty-dependent weight $1-\gamma_t$ with a constant branch weight $c\in\{0.2, 0.5, 1.0\}$ gives weaker performance across settings. Although $c=0.5$ is slightly more competitive, \Ours{} still achieves the highest average accuracy. This supports the main design intuition: teacher guidance should be applied more strongly when the observed token is uncertain, rather than uniformly across all tokens. Overall, the ablations validate both components of \Ours{}, showing that using a teacher-guided prior as a fallback mechanism in the residual branch is highly effective.

\newpage
\section{Teacher Model Alignment}
\label{app:teacher_alignment}

\begin{figure}[htbp]
  \centering
  \includegraphics[width=0.9\linewidth]{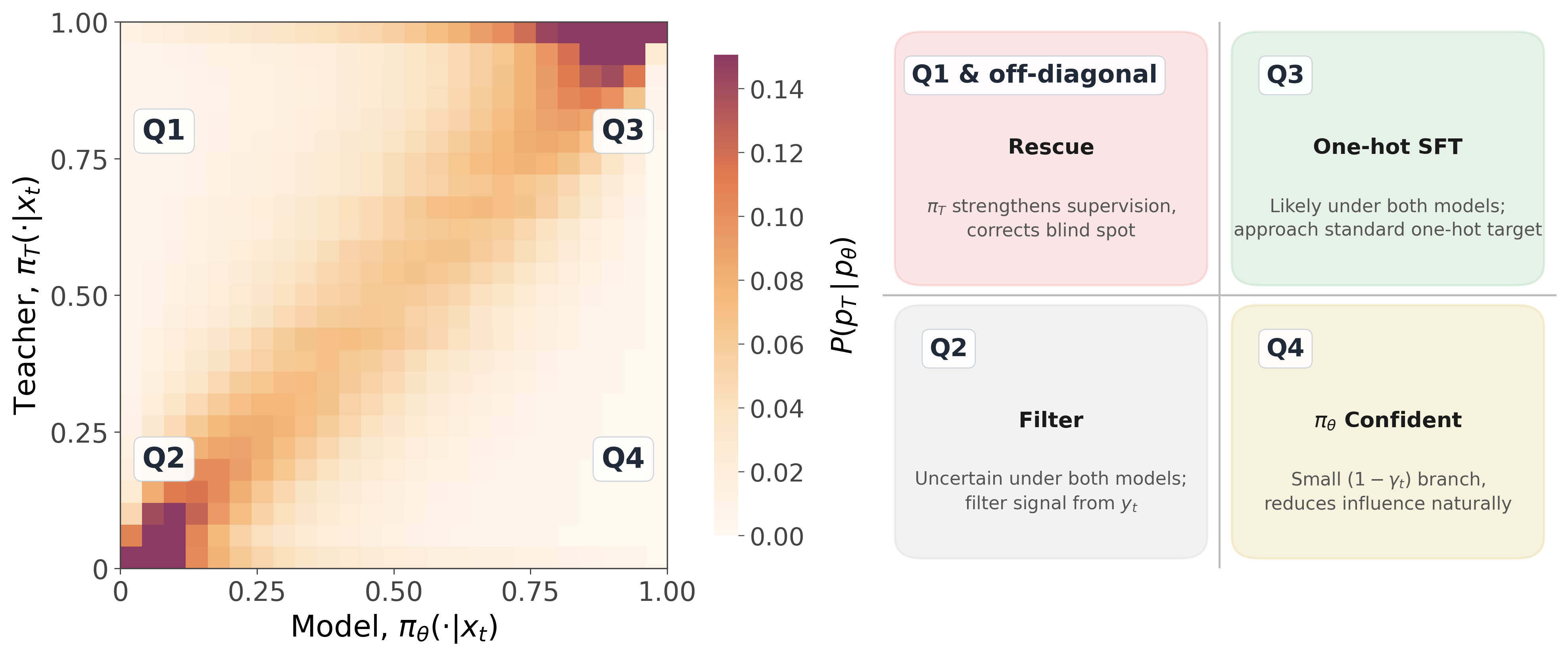}
  \caption{\textbf{Conditional Distribution of Probabilities.} This visualizes $P(p_T \mid p_\theta)$, the teacher probability $p_T$ given the student probability $p_\theta$ on the observed token $y_t$. Each column represents a fixed $p_\theta$ bin, with color intensity showing the empirical density of $p_T$ within that bin. The four annotated quadrants define qualitatively distinct supervision regimes.
}
  \label{fig:joint_density}
\end{figure}

This section analyzes the alignment the model $\pi_\theta$ and teacher distribution $\pi_T$ to better understand the effects of teacher guidance. Figure~\ref{fig:joint_density} visualizes the conditional distribution $P(p_T \mid p_\theta)$ of the teacher probability $p_T=\pi_T(\cdot \mid x_t)$ given the policy model probability $p_\theta=\pi_\theta(\cdot \mid x_t)$, both evaluated on the observed ground-truth training token $y_t$. 

A large portion of tokens lies near the diagonal, where both models assign similar probabilities to $y_t$. However, the distribution shows meaningful spread around the diagonal. This indicates useful correction signal, since even modest deviations from the diagonal are cases where the teacher's confidence differs and thus providing signals on alternative tokens in vocabulary. While the visualization only shows the marginal probabilities on $y_t$ for clarity, we note that the more important effect is the teacher's full redistribution of probability mass across alternative tokens. And divergence on $y_t$ serves as a proxy for for broader differences in the teacher's beliefs over the vocabulary, enabling fine-grained adjustments that the student's current distribution cannot capture.

The off-diagonal density in Q1 represents stronger disagreements (high $p_T$, low $p_\theta$), where teacher guidance is most informative. These tokens correspond to cases where the base model lacks confidence, but the teacher recognizes the token as plausible or important. This shows the role of teacher guidance as a fallback signal: when the student is uncertain, the teacher strengthened supervision rather than relying solely on the student’s current belief. Notably, many tokens cluster at $p_T \approx 1.0$, indicating that the teacher frequently provides strong corrections on tokens the student underweights.

In Q2 (both $p_T$ and $p_\theta$ low), both models are uncertain about the observed token. These may correspond to noisy, idiosyncratic, or highly non-unique tokens. Standard SFT treats them as fully reliable labels, forcing exact imitation through one-hot target $\delta_{y_t}$. In contrast, a softer target reduces imitation strength here, allowing probability mass to plausible alternatives. This helps avoid fitting dataset artifacts or arbitrary surface choices.

In Q3, the two models assign high probability to $y_t$. These are high-confidence tokens where the label is likely reliable and unambiguous. The target thus remains close to standard SFT one-hot $\delta_{y_t}$, since both models support strong imitation. Finally, in Q4 (high $p_\theta$, low $p_T$), the student model is already confident but the teacher is uncertain. In this case, the teacher supervision is downweighted through a smaller $1-\gamma_t$, naturally reducing its influence.

Together, this analysis supports the motivation of \Ours{}, providing a selective fallback distribution to adjust the confidence on dataset tokens. It strengthens supervision when the model is uncertain, while relaxing imitation on low-confidence uncertain tokens. This provides a more adaptive alternative to standard SFT, which imposes the strict one-hot target $\delta_{y_t}$ regardless of token uncertainty or model-data alignment. Appendix~\ref{app:ex_trajectories} provides example trajectories showing tokens in the Q1 (high $p_T$, low $p_\theta$) and Q2 (both low), ideally corresponding to useful knowledge vs. uncertain/noisy tokens.

\newpage
\section{Qualitative Examples}
\label{app:ex_trajectories}

\begin{figure}[htbp]
  \centering
  \includegraphics[width=0.96\linewidth]{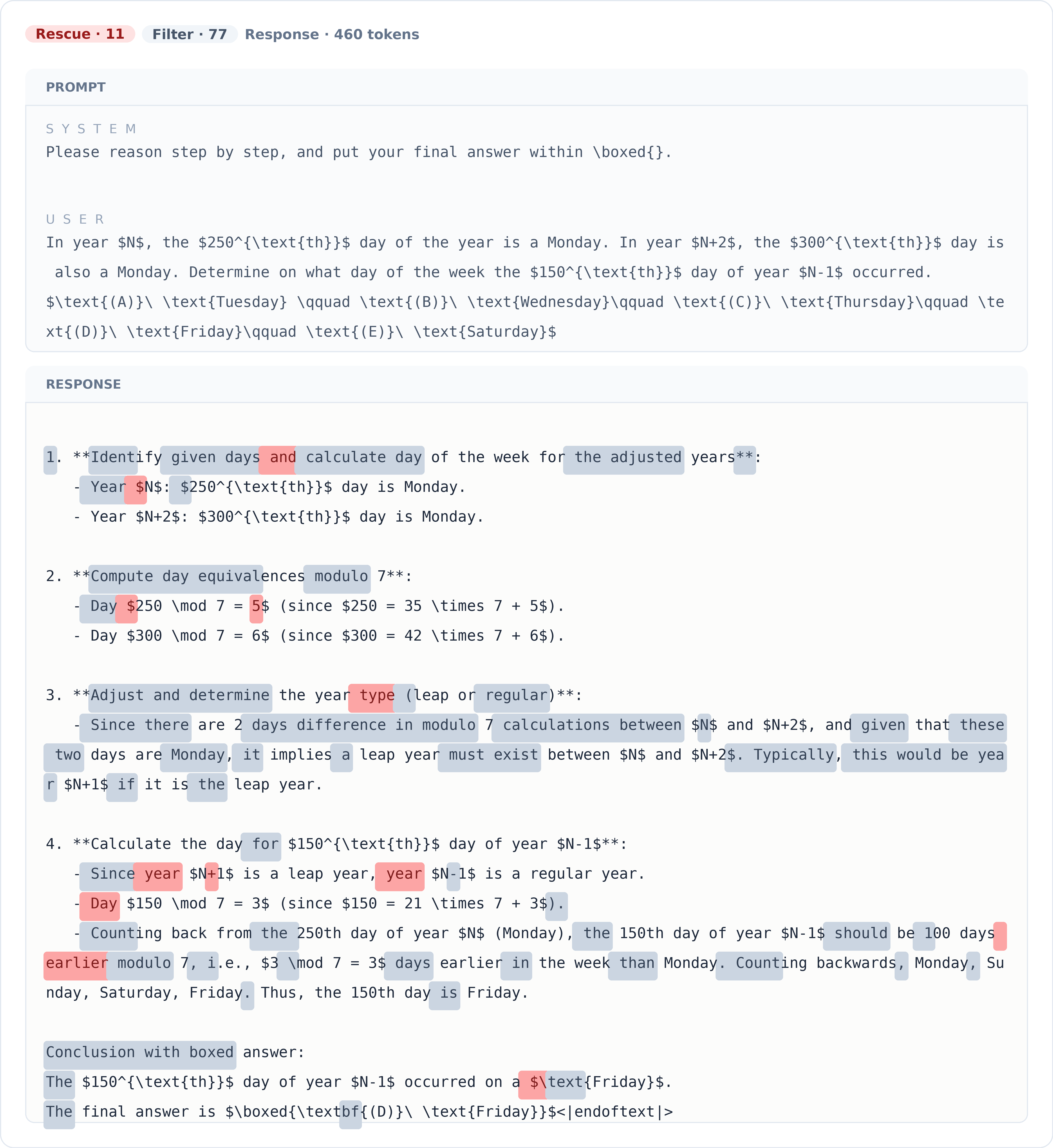}
  \caption{\textbf{Example Trajectory \#1.}}
  \label{fig:traj_2}
\end{figure}

\begin{figure}[htbp]
  \centering
  \includegraphics[width=\linewidth]{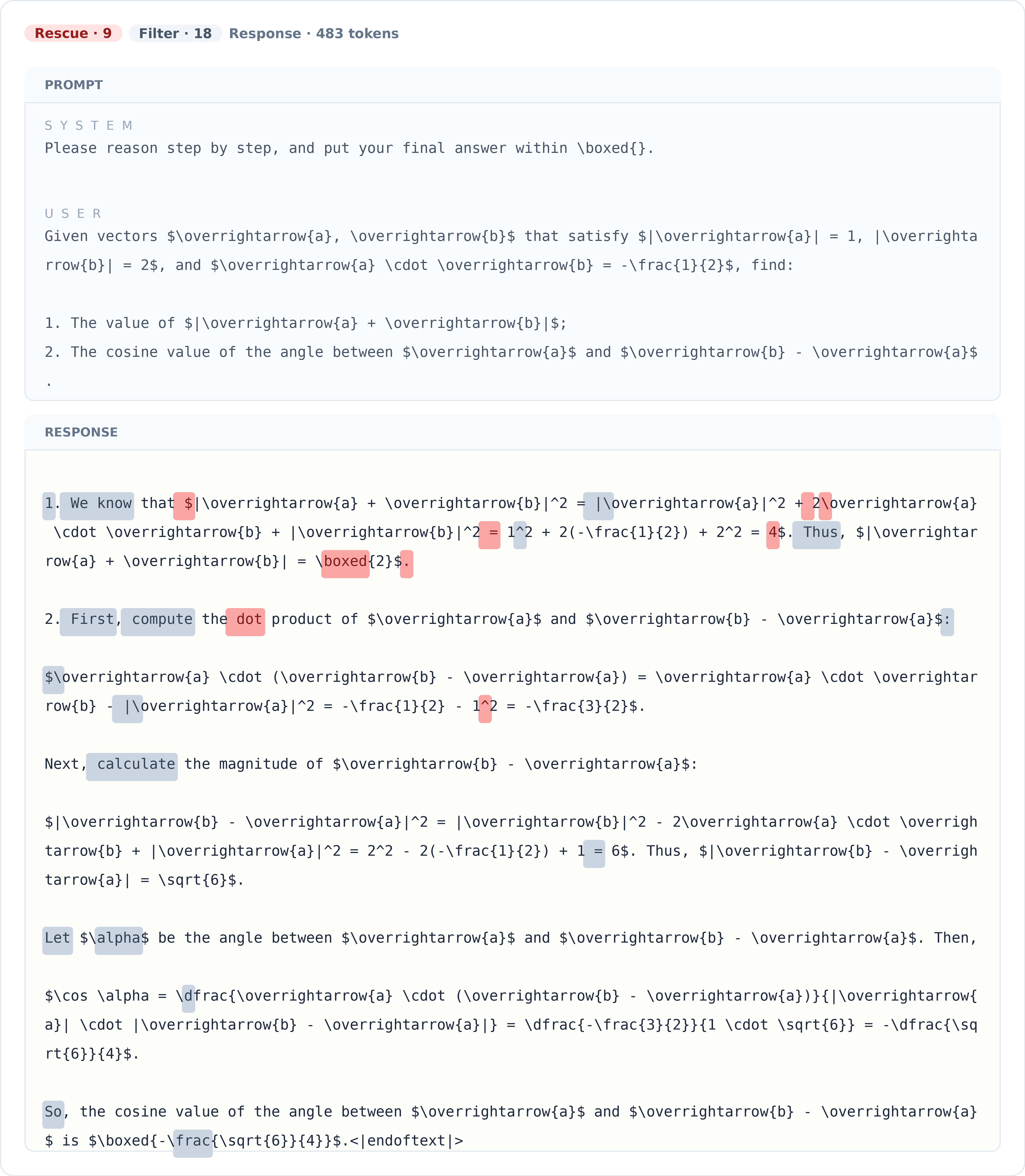}
  \caption{\textbf{Example Trajectory \#2.}}
  \label{fig:traj_1}
\end{figure}

Figures~\ref{fig:traj_1},~\ref{fig:traj_2}, ~\ref{fig:traj_3} show example trajectories from NuminaMath-67k dataset. {\color{red!60}\textbf{Rescue}} token $y_t$ is assigned a low probability under $\pi_\theta$ but high probability under $\pi_T$, hence strengthens the supervision that is otherwise weak. {\color{black!40}\textbf{Filter}} token $y_t$ has low probability under both $\pi_\theta$ and $\pi_T$, thus relaxing its imitation strength. Rescue tends to be answer-binding structural tokens, while Filter tends to be stylistic or chain-of-thought bridge words (e.g., \textit{1.}, \textit{First}, \textit{Thus}, \textit{Therefore}, \textit{So}, etc)

\begin{figure}[htbp]
  \centering
  \includegraphics[width=\linewidth]{img/appendix/example_216.png}
  \caption{\textbf{Example Trajectory \#3.}}
  \label{fig:traj_3}
\end{figure}

\newpage
\section{Response Length}

\begin{figure}[htbp]
  \centering
  \includegraphics[width=\linewidth]{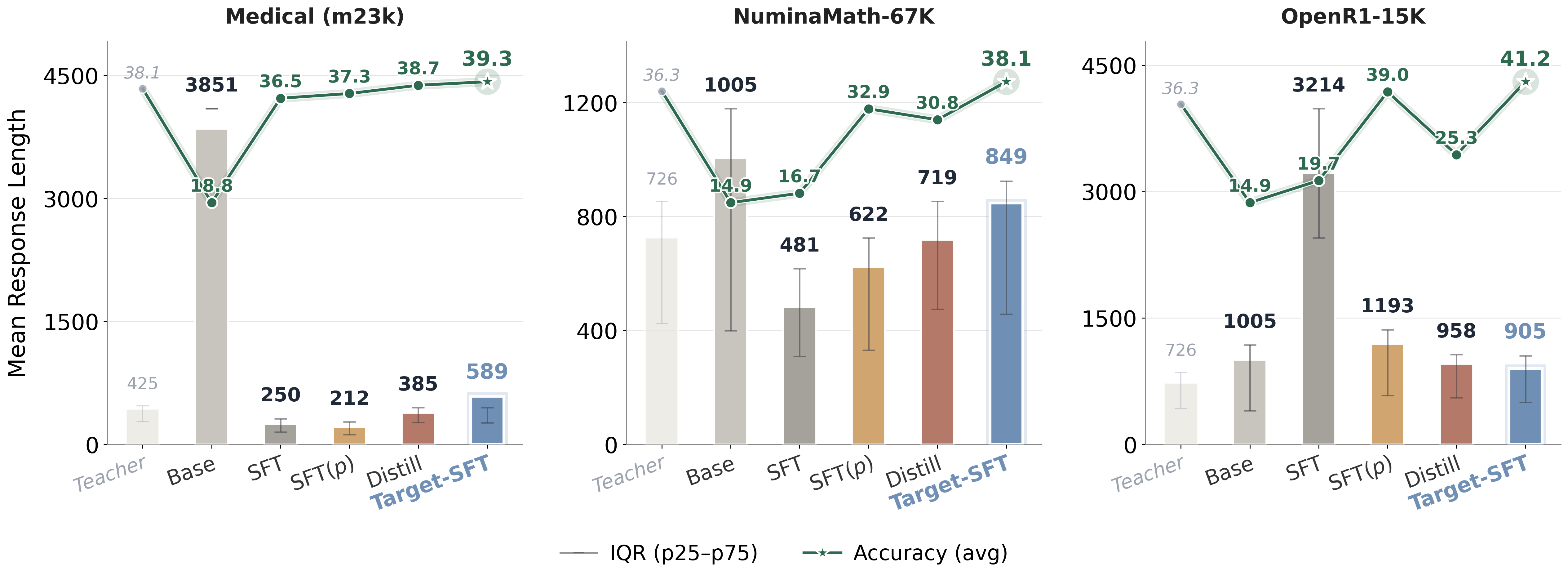}
  \caption{\textbf{Comparison of Response Length.} Bars indicate mean response length (tokens), error bars show the interquartile range (p25--p75), and the green curve reports average accuracy from evaluation in the main text. Response length does not consistently predict performance: long outputs from the base model or standard SFT often reflect rambling, repetition, or dataset-specific style imitation, while \Ours{} achieves strong accuracy with more moderate and stable response lengths.}
  \label{fig:length}
\end{figure}

This section analyzes the relationship between the model's output response length and its accuracy performance. Figure~\ref{fig:length} shows the comparison on Qwen2.5-Math-1.5B for the two mathematical reasoning datasets, and Qwen2.5-1.5B for the m23k dataset.

On m23k and NuminaMath-67K, trained models that produce longer responses generally have higher accuracies than the shorter-output SFT variants, suggesting that the increase in reasoning length is beneficial. However, this relationship between response length and model performance is not consistent on OpenR1-15K, where standard SFT produces the longest responses but achieves substantially lower accuracy than the others. Meanwhile, \Ours{} achieves the highest accuracy with relatively shorter responses. This indicates that this pattern may be dataset-dependent and response length alone is not a reliable proxy for reasoning quality.

A closer inspection of model outputs suggests that the long responses from the base model often reflect poor generation behavior rather than longer reasoning. In particular, the base model occasionally shows rambling, unstable formatting, repetition, or failure to terminate properly. This inflates the average response length while yielding weak performance. In contrast, all trained models produce more consistent outputs and more stable response formats.

Standard SFT appears especially sensitive to the surface style of the training corpus. For example, on OpenR1-15K, it tends to imitate the dataset’s long chain-of-thought monologue style, closely matching its response length and tone. This observation aligns with the intuition behind SFT's one-hot target $\delta_{y_t}$, which applies strict supervision to every observed token in the training data. As a result, standard SFT may overfit to dataset-specific surface forms, including verbosity and stylistic structures, without necessarily improving the robustness of the underlying reasoning process.

By comparison, SFT($p$), distillation, and \Ours{} tend to preserve more of the base model’s flexibility. Their outputs fall into a more moderate-length regime and avoid the extreme verbosity of standard SFT. These methods appear to trade some of SFT’s confident dataset-matched style for a more consistent reasoning structure that transfers better across corpora, leading to stronger accuracy without simply increasing response length or mimicking styles from the dataset.



\end{document}